\documentclass[journal]{IEEEtran}
\usepackage{subfig}
\usepackage{graphicx} 
\usepackage{amsmath}

\usepackage[linesnumbered,ruled,vlined]{algorithm2e}
\usepackage{algpseudocode}
\usepackage{float,amsthm}
\usepackage{amsfonts}
\usepackage{booktabs}

\usepackage{mathtools}

\usepackage{threeparttable}
\usepackage{multirow}

\usepackage{array}
\newcolumntype{P}[1]{>{\centering\arraybackslash}p{#1}}

\usepackage{cite}
\usepackage{copyrightbox}

\usepackage{tikz}
\newcommand*\circled[1]{\tikz[baseline=(char.base)]{\node[shape=circle,draw,inner sep=2pt] (char) {#1};}}

\makeatletter
\renewcommand{\CRB@setcopyrightfont}{%
\usefont{T1}{phv}{m}{n}\fontsize{8}{1}\selectfont
}
\makeatother

\theoremstyle{definition}

\newtheorem{definition}{Definition}

\newcounter{remark}

\newcounter{example}

\newcounter{property}

\usepackage{color}
\newcommand{\ParSection}[1]{}
\begin{document}

\twocolumn[
\begin{@twocolumnfalse}
\vspace*{5cm}

\copyright 2020 IEEE.  Personal use of this material is permitted.  Permission from IEEE must be obtained for all other uses, in any current or future media, including reprinting/republishing this material for advertising or promotional purposes, creating new collective works, for resale or redistribution to servers or lists, or reuse of any copyrighted component of this work in other works.
\end{@twocolumnfalse}
]
\newpage

\title{Extending the Morphological Hit-or-Miss Transform to Deep Neural Networks}

\author{Muhammad~Aminul~Islam,~\IEEEmembership{Member,~IEEE,}
      Bryce~Murray,~\IEEEmembership{Student Member,~IEEE,}
      Andrew~Buck,~\IEEEmembership{Member,~IEEE,}
      Derek~T.~Anderson,~\IEEEmembership{Senior~Member,~IEEE,}
      Grant~Scott,~\IEEEmembership{Senior~Member,~IEEE,}
       Mihail~Popescu,~\IEEEmembership{Senior~Member,~IEEE,}
       James~Keller,~\IEEEmembership{Life~Fellow,~IEEE,}% <-this % stops a space
\thanks{Muhammad Aminul Islam is with the Department of Electrical \& Computer Engineering and Computer Science, University of New Haven, Connecticut, CT 06516, USA. E-mail: (amin\_b99@yahoo.com).}% <-this % stops a space
\thanks{Bryce~Murray, Andrew~Buck, Derek~T.~Anderson, Grant~J.~Scott, Mihail~Popescu, and James~Keller are with the Department of Electrical Engineering and Computer Science, University of Missouri, Columbia, MO 65211.}% <-this % stops a space
\thanks{}% <-this % stops a space
\thanks{Manuscript revised June, 2020.}}

% The paper headers
%\markboth{Draft Submitted to the IEEE Transactions on Neural Networks and Learning Machines}%
%{Shell \MakeLowercase{\textit{et al.}}: Draft Submitted to the IEEE Transactions on Neural Networks}
\markboth{IEEE Transactions on Neural Networks and Learning Systems}%
{Islam \MakeLowercase{\textit{et al.}}: Draft Submitted to the IEEE Transactions on Neural Networks and Learning Systems}

%%DEREK%%%%%%%%%%%%%%%
%\tableofcontents
%\newpage
%%%%%%%%%%%%%%%%%%%%%%

\maketitle

\begin{abstract}
% Neural networks have demonstrated breakthrough results in numerous computer vision tasks. While most architectures  use convolution as a fundamental operation, recently there have been some works that explored morphology, an interpretable feature learning technique, that evolved from set theory and has long been used for analysis and processing of geometric structures.
% Herein, we explore the hit-or-miss transform, a morphological operation, that takes into account both the foreground and background in measuring the fitness of a target shape in an image. 
% We identify limitations of the current hit-or-miss definition and formulate an optimization problem to learn the transform appropriately. Based on this formulation, we propose an algorithm to learn the  transform in the context of neural network.
% Our analysis shows that convolution, in fact, acts as a hit-miss transform, though the semantic meaning of its filters differs. Analogous to the generalized hit-or-miss transform, we introduce an extension of convolution that is demonstrated to outperform standard convolution on benchmark datasets. We conducted experiments on synthetic and benchmark datasets that show
% that hit-or-miss transform provides better interpretability with learned shapes consistent with input objects while convolution yields higher classification accuracy.

While most deep learning architectures are built on convolution, alternative foundations like morphology are being explored for purposes like interpretability and its connection to the analysis and processing of geometric structures. The morphological hit-or-miss operation has the advantage that it takes into account both foreground and background information when evaluating target shape in an image. Herein, we identify limitations in existing hit-or-miss neural definitions and we formulate an optimization problem to learn the transform relative to deeper architectures. To this end, we model the semantically important condition that the intersection of the hit and miss structuring elements (SEs) should be empty and we present a way to express Don't Care (DNC), which is important for denoting regions of an SE that are not relevant to detecting a target pattern. Our analysis shows that convolution, in fact, acts like a hit-miss transform through semantic interpretation of its filter differences. On these premises, we introduce an extension that outperforms conventional convolution on benchmark data. Quantitative experiments are provided on synthetic and benchmark data, showing that the direct encoding hit-or-miss transform provides better interpretability on learned shapes consistent with objects whereas our morphologically inspired generalized convolution yields higher classification accuracy. Last, qualitative hit and miss filter visualizations are provided relative to single morphological layer.

% (1) Discovered that CNN can already more-or-less do hit-and-hiss
% (2) But, if one wants to go morph route, we outline the "right way" to do it
% (3) End of day, doing this, get most explainable outline as possible
% (4) Also strong evidence that deep MSNN is not a good idea

\end{abstract}


\IEEEpeerreviewmaketitle

\section{Introduction}

\ParSection{Why Care?} 
% a.	Introductory paragraph: Discuss applications and importance of deep learning

% Deep learning has demonstrated a strong predictive performance across a wide variety of settings.

Deep learning has demonstrated robust predictive accuracy across a wide range of applications. Notably, it has achieved and, in some cases, surpassed human-level performance in many cognitive tasks, for example, object classification, detection, and recognition, semantic and instance segmentation, and depth prediction.
This success can be attributed in part to the ability of a \textit{neural network} (NN) to construct an arbitrary and very complex function by composition of simple functions, thus empowering it as a formidable machine learning tool.

To date, state-of-the-art deep learning algorithms mostly
use convolution as their fundamental operation, thus the name \textit{convolutional neural network} (CNN). Convolution has a rich and proud history in signal/image processing, for example extracting low-level features like edges, noise filtering (low/high pass filters), frequency-orientation filtering via the Gabor, etc. In a continuous space, it is defined as the integral of two functions---an image and a filter in the context of image processing---after one is reversed and shifted, whereas in discrete space, the integral realized via summation. CNNs progressively learn more complex features in deeper layers with low level features such as edges in the earlier layers and more complex shapes in the later layers, which are composite of features in the previous layer. While that has been the claim of many to date, recent work has emerged suggesting that mainstream CNNs--e.g., GoogLeNet, VGG, ResNet, etc.--are not sufficiently learning to exploit shape. In \cite{geirhos2018imagenettrained}, Geirhos et al. showed that CNNs are strongly biased towards recognizing texture over shape, which as they put it ``is in stark contrast to human behavioural evidence and reveals fundamentally different classification strategies.'' Geirhos et al. support these claims using a total of nine experiments totaling 48,560 psychophysical trials with respect to 97 observers. Their research highlights the gap and stresses the importance of shape as a central feature in vision.

An argument against convolution is that its filter does not lend itself to interpretable shape. Because convolution is correlation with a time/spatial reversed filter, the filter weights do not necessarily indicate the absolute intensities/levels in shape. Instead, they signify relative importance. Recently, investigations like guided backpropagation \cite{springenberg2014striving} and saliency mapping \cite{simonyan2013deep} have made it possible to visualize what CNNs are perhaps \emph{looking} at. However, these algorithms are not guarantees, they inform us what spatial locations are of interest, not what exact shape, texture, color, contrast, or other features led a machine to make the decision it did. Furthermore, these explanations depend on an input image and the learned filters. The filters alone do not explain the learned model. In many applications, it is not important that we understand the chain of evidence that led to a decision. The only consideration is if an AI can perform as well, if not better, than a human counterpart. However, other applications, e.g., medical image segmentation in healthcare or automatic target recognition in security and defense, require glass versus black box AI when the systems that they impact intersect human lives. In scenarios like these, it is important that we ensure that shape, when/where applicable, is driving decision making. Furthermore, the ability to seed, or at a minimum understand what shape drove a machine to make its decision is essential.

In contrast to convolution, morphology-based operations are more interpretable---a property well-known and well-studied in image processing, which has only been lightly studied and explored in the context of deep neural networks \cite{mellouli2019morphological,halkiotis2007automatic,won1995morphological,won1997morphological,zheng2006morphological,sulehria2008vehicle,jin2007vehicle,raducanu2001morphological,gader2000morphological,hocaoglu2003domain,khabou2000ladar,theera1998detection,ouadou2017vehicle}. Morphology is based on set theory, lattice theory, topology and random functions and has been used for the analysis and processing of geometric structures \cite{perret2009robust,chatzis2000generalized,ta2010nonlocal,bouaynaya2008theoretical,ji1992fast,zana2001segmentation,urbach2007connected,palmer1997locating,haralick1987image,sinha1992fuzzy}. The most fundamental morphological operations are erosion and dilation, which can be combined to build more complex operations like opening, closing, the hit-or-miss transform, etc. Grayscale erosion and dilation are used to find the minimal offset by which the foreground and background of a target pattern fits in an image, providing an absolute measure of fitness in contrast to relative measure by convolution, facilitating the learning of interpretable structuring elements (SEs).

Recently, a few deep neural networks have emerged based on morphological operations like dilation, erosion, opening, and closing \cite{mellouli2019morphological,nogueira2019introduction}. In \cite{mellouli2019morphological}, Mellouli et al. explored pseudo-dilation and pseudo-erosion defined in terms of an weighted counter harmonic mean, which can be carried out as the ratio of two convolution operations. However, their network is not an end-to-end morphological network, rather a hybrid of traditional convolution and pseudo-morphological operations. In \cite{nogueira2019introduction}, Nogueira et al. proposed a neural network based on binary SEs (consisting of $1$s and $0$s) indicating which pixels are relevant to the target pattern. Their proposed implementation requires a large number of parameters, specifically $s^2$ binary filters of size $s\times s$ just to represent a single $s\times s$ SE; making the method expensive in terms of storage and computation and not suitable for deep learning.  Furthermore, they did not conduct any experiments nor provide results for popular computer vision benchmark datasets, e.g., MNIST or Cifar. More importantly, none of these algorithms simultaneously apply dilation and erosion on an image to take into account both foreground and background. In the morphological community, there is a well-known operation for achieving this, the hit-miss transform, the subject of our current article.

Following the success of convolution based shared weight neural networks on handwritten digit recognition tasks, Gader et al. introduced a generalized hit-or-miss transform network, referred to as image algebra network \cite{gader1994image}. Later, the standard hit-or-miss transform was applied in target detection \cite{won1997morphological}. All of these methods employed two SEs, one for the hit to find the ``fitness'' of an image relative to target foreground and another for the miss to find the ``fitness'' relative to target background. However, existing grayscale hit-or-miss transform definitions as well as their neural network implementations \cite{gader1994image,khabou1999morphological,won1995morphological,perret2009robust,dougherty1992introduction,gonzalez2002digital} neither state nor enforce the condition that the intersection of the hit and miss SEs must be empty. Failing to meet this condition can result in semantically inconsistent and uninterpretable SEs. To address this, we put forth an optimization problem enforcing the non-intersecting condition.

However, considering only foreground and background are not sufficient to describe target shape. We also need \textit{Don't Care} (DNC), which denotes regions of the SE that are not relevant to detecting a target pattern. While binary morphology considers 0s as DNCs and it ignores them during computation, it's grayscale extension unfortunately considers all elements including 0s. Therefore, we propose a new extension to the hit-or-miss transform which allows it to describe a grayscale shape in terms of relevant and non-relevant elements (i.e., DNC). Herein, we provide the conditions that will make elements under the conventional definition of hit-or-miss to act as DNC and we show that the valid ranges for target and DNC elements are discontinuous. However, this constraint poses a challenge to data-driven learning using gradient descent, which requires the variables to reside in a (constrained or unconstrained) continuous space. As a result, we propose hit-or-miss transforms that implicitly enforces non-intersecting condition and addresses DNC. 

Last, while convolution can act like a hit-or-miss transform -- when its ``positive filter weights'' correspond to foreground, ``negative weights'' to background, and 0s for DNC -- it differs in some important aspects. For example, elements in a hit-or-miss SE indicate the absolute intensity levels in the target shape whereas weights in a convolution filter indicate relative levels/importance. Another difference is that the sum operation gives equal importance to all operands versus max (or min) in the hit-or-miss. On this premises, we propose a new extension to convolution, referred to as generalized convolution hereafter, by replacing the sum with the generalized mean. The use of a parametric generalized mean allows one to choose how values in the local neighborhood contribute to the result; e.g., all contribute equally (as in the case of the mean) or just one drives the result (as in max), or something in between. Through appropriate selection of this parameter, performance can be significantly enhanced as demonstrated by our experiments. 

While convolution, likewise the hit-or-miss transform, consider foreground, background, and DNC, they differ in how fitness is evaluated. For example, convolution uses a relative measure whereas the hit-or-miss uses an absolute measure. One question naturally arises, how does this difference impact performance on two aspects of a learned model, explainability and accuracy. Our analysis (Section \ref{sec:exp}) shows that morphology provides better interpretability through its use of an absolute measure, while convolution yields higher accuracy as a relative measure is more robust. 

In summary, our article makes the following specific contributions to neural morphology.
\begin{itemize}
    \item We identify limitations in the current definition of the grayscale hit-or-miss and we formulate an optimization to properly learn the transform in a neural network.
    \item In light of this optimization, we propose an algorithm to learn the hit-or-miss transform and also its generalization.
    \item We extend ``conventional convolution'' used in most neural networks with a parametric generalized mean.
    \item Synthetic and benchmark datasets are used to show the behavior and effectiveness of the proposed theories in a quantitative (via accuracy) and qualitative (via preliminary shallow, single layer, filter visualizations) respect.
\end{itemize}

The remainder of this article is organized as follows. In Section \ref{sec:background}, we provide notations and definitions of binary and grayscale morphological operations. Section III introduces the optimization problem, learning algorithm, our generalization of the hit-or-miss transform, and our extension of convolution, followed by experiments and results in Section IV.

\section{Binary and Grayscale Morphology}\label{sec:background}

First, we briefly review definitions related to binary morphology in order to understand the basis of morphological operations and their semantic meaning pertaining to image processing. The most basic of morphological operations are dilation and erosion, which coupled with algebraic operations (e.g., sum) create more complex morphological operations like opening, closing, hit-or-miss transform, top-hat, thinning, thickening, and skeleton, to name a few.

\subsection{Binary Morphology}

Binary morphology is grounded in the theory of sets. Let $Z$ be a set of integers.
\begin{definition} (\textbf{Dilation}) Let $A$ be an image, $B$ a SE, and $A, B \in Z^2$. The dilation of $A$ by $B$, denoted by $A \oplus B$ is
\[A \oplus B = \{z| (\hat{B})_z \cap A \neq \emptyset\},\]
where $\hat{B}$ is the reflection of $B$ about its origin and $(B)_z$ is the translation of $B$ by $z$ \cite{gonzalez2002digital,dougherty1992introduction}. 
\end{definition}
As the above definition shows, the dilation operation involves reflecting $B$ and then shifting the reflected $B$ by $z$. The dilation of $A$ by $B$ is the set of all displacements $z$ such that $B$ and $A$ overlap by at least one element. 
The set $B$ is often referred to as the \textit{structuring element} (SE).

\begin{definition} (\textbf{Erosion}) Let $A$ be an image, $B$ a SE, and $A, B \in Z^2$. Then the erosion of $A$ by $B$, denoted $A \ominus B$ is
\[A \ominus B = \{z| (B)_z \subseteq A\},\]
where $(B)_z$ is translation of $B$ by $z$ \cite{gonzalez2002digital,dougherty1992introduction}.
\end{definition}
The above equation indicates that $A \ominus B$ is the set of all points such that $B$, translated by $z$, is contained in $A$. 

%SEs can be ``non-flat'', when the weights are non-uniform or flat when the weights are uniform. With ``flat'' SEs, particularly when elements are zeros, dilation is the maximum of the image pixels values in the window erosion is the minimum in the window. Thus, dilation and erosion become order statistics with flat structuring elements. 
% Many applications, e.g., ..., use flat structuring elements.

It is a well-known fact that dilation and erosion are duals of each other with respect to complement and reflection.
\[(A\ominus B)^c = A^c \oplus \hat{B},\]
where $A^c$ is the complement of $A$. Similarly,
\[(A\oplus B)^c = A^c \ominus \hat{B}.\]

The morphological hit-or-miss transform is a technique for shape detection that simultaneously matches both foreground and background shapes in an image.
\begin{definition}\textbf{(Binary Hit-or-Miss)}
The binary hit-or-miss transform w.r.t. SEs $H$ and $M$ satisfying $H\cap M = \emptyset$ is
\[A \odot (H,M)= (A\ominus H) \cap (A^c \ominus M),\]
where $H$ is the set associated with the foreground or an object and $M$ is the set of elements associated with the background.
\label{def:binaryHitMiss}
\end{definition}

The elements in a binary SE are indexed w.r.t the origin (or a reference point) that can be designated to any point within the SE.
%but the top-left corner in accordance with the conventional image co-ordinate system has mostly been used \cite{}.
In order to compute the transform, both $H$ and $M$ are slided over the binary image for every possible locations.
In this way, $A \odot (H,M)$ finds all the points (origins of the translated structuring elements) at which, simultaneously, $H$ found a match (``hit") in $A$ and $M$ found a match in $A^c$. By using the dual relationship between erosion and dilation, the hit-or-miss transform equation can alternatively be written as
\begin{equation}
    A \odot B = (A\ominus H) \setminus (A\oplus \hat{M}),
\label{eq:binaryHitMiss}
\end{equation}
where $\setminus$ is the set difference operation ($A\setminus B = A\cap B^c$).

Though obvious from Def.~\ref{def:binaryHitMiss}, we emphasize that the intersection of sets that define the foreground (aka hit) and background (aka miss) must be null or empty, i.e., both $H(x,y)$ and $M(x,y)$ at a given location $(x,y)$ cannot be $1$. This is because an element in the target structure can either be treated as foreground, background or DNC (an element not part of the target structure and is defined by $0$s in both hit and miss SEs) but it cannot simultaneously be foreground and background.
% as it is counter-intuitive, contradictory and logically impossible. 
% Therefore, if a pixel is $1$ in hit SE, it must be $0$ in miss SE and vice versa, otherwise the transform will always result in an empty set. 
We illustrate all these cases (e.g., non-intersecting and intersecting SEs) with examples in Fig.~(\ref{fig:binaryMorphExample}). Table~\ref{tab:binaryMorph} shows combination of  hit-miss values for binary morphology.

\begin{table}[htbp]
  \centering
  \renewcommand{\arraystretch}{1.5}
  \caption{Binary combinations for the hit-or-miss transform in binary morphology}
    \begin{tabular}{ccc}
    \toprule
    H & M & Semantic meaning \\
    % Corresponding element in the target pattern
    \midrule
    0 & 0 & DNC\\
    0 & 1 & Background\\
    1 & 0 & Foreground\\
    1 & 1 & Inadmissible - semantically infeasible\\
    \bottomrule
    \end{tabular}%
  \label{tab:binaryMorph}%
\end{table}%

\begin{figure*}
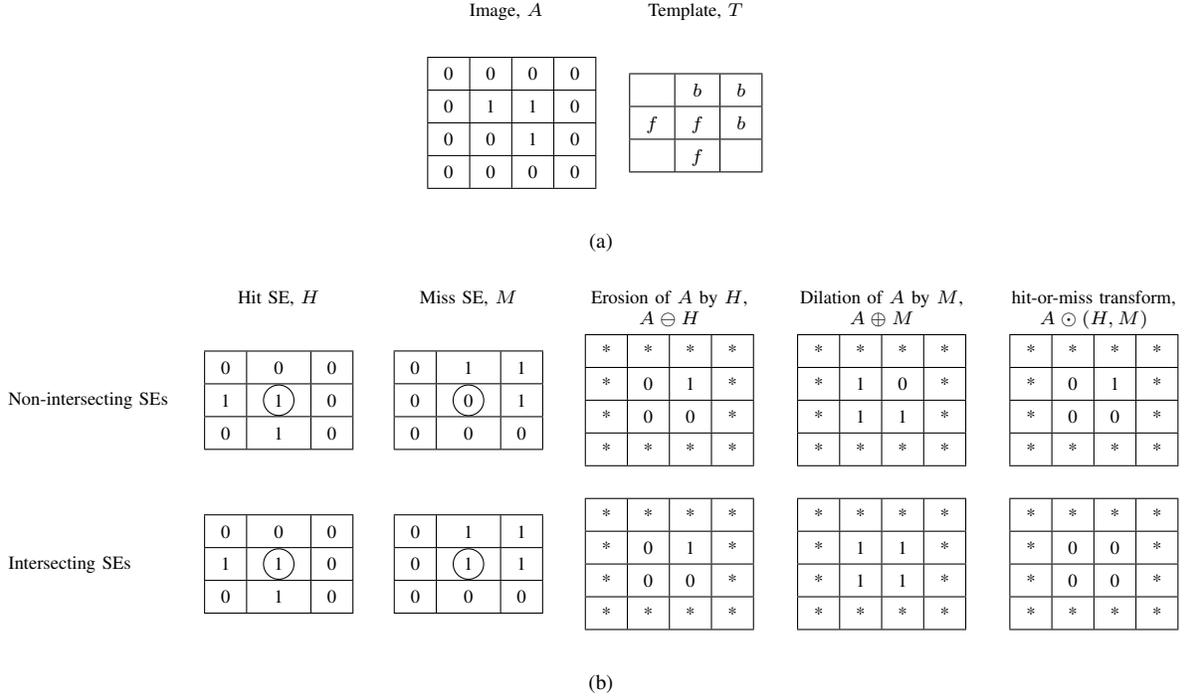

\scriptsize
%     \centering
%     \caption{Caption}
% \begin{table*}[htbp]
  \centering
  \renewcommand{\arraystretch}{1.5}
  \subfloat[]{
     \begin{tabular}{*{2}{P{2.1cm}}}
     Image, $A$ & Template, $T$\\
     \\
     \begin{tabular}{|*{4}{c|}}
\hline
    0 & 0 & 0 & 0  \\
    \hline
    0 & 1 & 1 & 0 \\
    \hline
    0 & 0 & 1 & 0  \\
    \hline
    0 & 0 & 0 & 0 \\
    \hline
\end{tabular}

&
\begin{tabular}{|*{3}{c|}}
\hline
     & $b$ & $b$  \\
     \hline
    $f$ & $f$  & $b$\\
    \hline
     & $f$ & \\
     \hline
\end{tabular}
     
     \end{tabular}
  }

  \subfloat[]{
    \begin{tabular}{@{}l *{2}{P{2.1cm}} *{3}{P{2.4cm}}}
    % \toprule
     & Hit SE, $H$ & Miss SE, $M$ & Erosion of $A$ by $H$, $A\ominus H$ & Dilation of $A$ by $M$, $A\oplus M$ & hit-or-miss transform, $A\odot (H,M)$\\
Non-intersecting SEs

&

\begin{tabular}{|*{3}{c|}}
\hline
    0 & 0 & 0  \\
\hline
    1 & \circled{1}  & 0\\
\hline
    0 & 1 & 0 \\
\hline
\end{tabular}

&

\begin{tabular}{|*{3}{c|}}
\hline
    0 & 1 & 1  \\
    \hline
    0 & \circled{0}  & 1\\
    \hline
    0 & 0 & 0 \\
    \hline
\end{tabular}

&
\begin{tabular}{|*{4}{c|}}
\hline
 * & * & * & *\\
 \hline
 * &   0 & 1 & *  \\
    \hline
* &  0 & 0 & *\\
\hline
 * & * & * & *\\
    \hline
\end{tabular}

&
\begin{tabular}{|*{4}{c|}}
\hline
 * & * & * & *\\
 \hline
 * &   1 & 0 & *  \\
    \hline
* &  1 & 1 & *\\
\hline
 * & * & * & *\\
    \hline
\end{tabular}

&
\begin{tabular}{|*{4}{c|}}
\hline
 * & * & * & *\\
 \hline
 * &   0 & 1 & *  \\
    \hline
* &  0 & 0 & *\\
\hline
 * & * & * & *\\
    \hline
\end{tabular}
\\
Intersecting SEs
&

\begin{tabular}{|*{3}{c|}}
\hline
    0 & 0 & 0  \\
\hline
    1 & \circled{1}  & 0\\
\hline
    0 & 1 & 0 \\
\hline
\end{tabular}

&

\begin{tabular}{|*{3}{c|}}
\hline
    0 & 1 & 1  \\
    \hline
    0 & \circled{1}  & 1\\
    \hline
    0 & 0 & 0 \\
    \hline
\end{tabular}

&
\begin{tabular}{|*{4}{c|}}
\hline
 * & * & * & *\\
 \hline
 * &   0 & 1 & *  \\
    \hline
* &  0 & 0 & *\\
\hline
 * & * & * & *\\
    \hline
\end{tabular}

&
\begin{tabular}{|*{4}{c|}}
\hline
 * & * & * & *\\
 \hline
 * &   1 & 1 & *  \\
    \hline
* &  1 & 1 & *\\
\hline
 * & * & * & *\\
    \hline
\end{tabular}

&
\begin{tabular}{|*{4}{c|}}
\hline
 * & * & * & *\\
 \hline
 * &   0 & 0 & *  \\
    \hline
* &  0 & 0 & *\\
\hline
 * & * & * & *\\
    \hline
\end{tabular}
\\
    % \bottomrule
    \end{tabular}%
    }
     \caption{Example of binary hit-or-miss transform structuring elements to detect a top-right corner. (a) shows a binary image, $A$, with a top-right corner in the top-right $3\times 3$ window and a $3\times 3$ template, $T$, that encodes the structure of the top-right corner and is used to construct SEs for hit-or-miss transform, $f$ stands for foreground, $b$ for background and empty cells are DNC. (b) Top row shows hit-or-miss transform for non-intersecting SEs derived from $T$, which correctly finds matching for both foreground in hit ($1$ in erosion means the foreground is matched) and background in miss ($0$ in dilation means the background is matched). Bottom row shows intersecting SEs, which produces empty set as it cannot find a matching for both foreground and background. Note that the transform is calculated without padding of the input image so the output size is 2 by 2. Note that we considered centers of the SEs as the origins, which are marked with circles.} 
\label{fig:binaryMorphExample}
\end{figure*}

\subsection{Grayscale Morphology}
% The morphological operations for binary images can be extended to grayscale image as well with  their definition as follows.
Let $f$ be a grayscale image, $b$ a structuring element, and $f(x,y)$ the grayscale intensity at a location $(x,y)$.
\begin{definition} {(\textbf{Grayscale Dilation})}   The grayscale dilation of $f$ by $b$, denoted as $f \oplus b$, is  \cite{gonzalez2002digital}
\begin{align}
    (f \oplus b) (x,y) = & \max \{f(s-x, t - y) + b(x,y) | \nonumber \\ & (s-x), (t-y) \in D_f;  (x,y) \in D_b\}, 
    \label{eq:grayscale-dilation}
\end{align}
where $D_f$ and $D_b$ are the domains of $f$ and $b$, respectively.
\end{definition}
% There is a parallel between 2-D convolution and dilation with the sum operation replacing the product and the max operation replacing sum of convolutions.
%On a side note, there is a parallel between 2-D convolution and dilation, when sum replaces product and when max replaces sum.

\begin{definition} {(\textbf{Grayscale Erosion})}  The grayscale erosion of $f$ by $b$, denoted as $f \ominus b$, is defined as 
\begin{align}
    (f \ominus b)  (x,y)= & \min \{f(s+x, t + y) - b(x,y) | \nonumber \\ & (s+x), (t+y) \in D_f;  (x,y) \in D_b\},
    \label{eq:grayscale-erosion}
\end{align}
where $D_f$ and $D_b$ are the respective domains \cite{gonzalez2002digital}.
\end{definition}
%\begin{remark}
%On a side note, there is a parallel between 2D correlation and erosion, when sum replaces product and when min replaces sum.}
%\end{remark}

As noted in \cite{dougherty1992introduction,gader1994image}, the umbra transform provides the theoretical basis for grayscale extension for morphological operation by providing a mechanism to express grayscale operations in terms of binary operations. Interested readers can refer to \cite{dougherty1992introduction,gader1994image} for the theory and proof of the extension.

A major difference between binary and grayscale morphology is that unlike binary morphological operations, there is no explicit DNC conditions in grayscale morphology, i.e., all elements including those with $0$s contribute to the results. So, a mechanism needs to be put in place to distinguish between target pixels and DNC. Ideally, the DNC elements can be specified by $-\infty$, which would result in maximum value for the erosion and minimum value for the dilation and thus will never contribute to the result. While suitable for hand-crafted SE design, it might not be feasible to learn $-\infty$-valued elements in the context of data-driven learning unless some constraints are imposed. Instead, the SEs can be designed such that the DNC elements are set to very low in compared to neighborhood elements so that the difference is always relatively high and as such it never carries over to the result. Thus, the filters can be designed smartly so that DNC is automatically enforced via appropriate selection of values. Alternatively, the erosion equation can be rewritten to consider only the foreground elements as in binary morphology.

Next, we find the condition for an element in an erosion SE to act as DNC. Let $I$ be an image in the interval $[lb_I, ub_I]$. Furthermore, let $h$ be the erosion SE with foreground elements in the interval $[lb_{h_f},ub_{h_f}]$ and DNC elements in the interval, $[lb_{h_d}, ub_{h_d}]$. Note that $lb_I$, $lb_{h_f}$, and $lb_{h_d}$ denote the lower bounds of image $I$, foreground elements in $h$, DNC elements in $h$, respectively. Similarly $ub_I$, $ub_{h_f}$, and $ub_{h_d}$ correspond to the respective upper bounds.  The maximum difference possible for foreground is $v_{max} = ub_I - lb_{h_f}$. We want the difference produced by DNC to be higher than $v_{max}$ for the lowest image value, $lb_I$ so that they are always ignored during the computation of $\max$. This leads to the condition, $lb_I- d \ge v_{max}$ or $d\le lb_I - ub_I + lb_{h_f}$, where $d$ is a DNC element. Consequently, $ub_{h_d}= lb_I - ub_I + lb_{h_f}$ and $lb_{h_d} = -\infty$. Since $lb_I - ub_I < 0$ for a grayscale image, $ub_{h_d} < lb_{h_f}$. This reveals that there is a discontinuity between valid ranges for foreground and DNC elements, i.e.,  a separation of $ub_I - lb_I$ must exist between them. This poses a challenge to the data-driven learning tasks since the weights learned are real-valued in a continuous domain and discontinuity cannot be enforced. Similar analysis can be performed for dilation SE, which yields the following condition, $lb_{m_d} < lb_{m_b}$, where $lb_{m_d}$ and $lb_{m_b}$ denote the lower bounds of background elements and DNC, respectively, in dilation SE $m$.

The hit-or-miss transform for grayscale is defined in literature via Eq.~(\ref{eq:binaryHitMiss}), by replacing the set difference operation with an arithmetic subtraction operation \cite{gonzalez2002digital, dougherty1992introduction,gader1994image}. 

\begin{definition} {(\textbf{Grayscale Hit-or-Miss Transform})}
The grayscale hit-or-miss transform is
\[f \odot (h,m) = (f \ominus h) - (f \oplus m^r),\]
where $m^r$ is the reflection of $m$, i.e., $m^r(x,y) = m(-x,-y)$, which gives
\begin{align*}
    (f & \odot (h,m))(x,y) =  \min \left\{(f(x+a,y+b) - h(a,b)) | \right. \\ & \left. (x+a), (y+b) \in D_f ; a,b\in D_h \right\} \\ & -  \max_{a,b\in D_m} \left\{ (f(x+a,y+b) + m(a,b)) \right. \\& \left. (x+a), (y+b) \in D_f ; a,b\in D_m \right\},
\end{align*}
where $(x+a), (y+b) \in D_f$ and  $D_f$, $D_h$, and $D_m$ are the domains of $f$, $h$, and $m$, respectively \cite{gonzalez2002digital}.
\end{definition}
% \begin{align*}
%     (f \odot (h,m))(x,y) = \min_{(x+a), (y+b) \in D_f ; a,b\in D_h} (f(x+a,y+b) - h(a,b)) \\ -  \max_{a,b\in D_m} (f(x+a,y+b) + m(a,b)),
% \end{align*}
% where $(x+a), (y+b) \in D_f$ and  $D_f$, $D_h$, and $D_m$ are the domains of $f$, $h$, and $m$, respectively \cite{gonzalez2002digital}.
% \end{definition}
Let $h$ and $m$ be SEs with non-negative weights. The hit and miss SEs together define the target pattern with hit indicating the foreground and miss indicating the background. For example, if $h(x,y)>m(x,y)$ - then that pixel is treated more as foreground than background and vice versa. 

\begin{figure*}
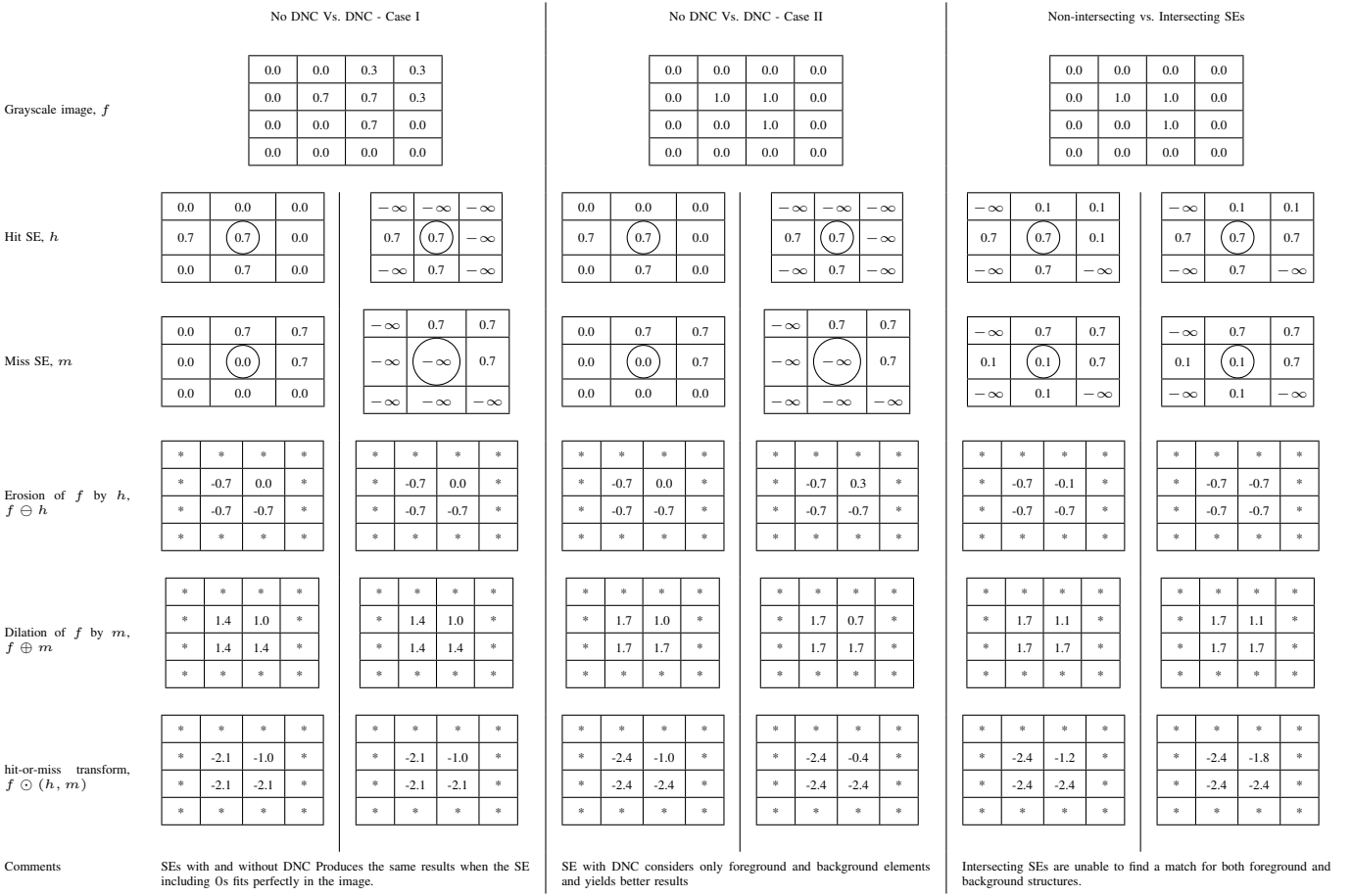

\tiny
%     \centering
%     \caption{Caption}
% \begin{table*}[htbp]
  \centering
  \renewcommand{\arraystretch}{1.7}
    % \begin{tabular}{@{}p{1.5cm} *{2}{P{2.1cm}} |*{2}{P{2.1cm}}| *{2}{P{2.1cm}}}
    % \resizebox{\textwidth}{!}{%
    \begin{tabular}{@{}p{1.7cm}  P{2.2cm} *{5}{|P{2.2cm}}}
    % \toprule
    
    & \multicolumn{2}{c|}{No DNC Vs. DNC - Case I} & \multicolumn{2}{c|}{No DNC Vs. DNC - Case II}& \multicolumn{2}{c}{Non-intersecting vs. Intersecting SEs} 
    \\
    & \multicolumn{2}{c|}{} & \multicolumn{2}{c|}{} & \multicolumn{2}{c}{}
    \\
    Grayscale image, $f$ 
    
    &
    
    \multicolumn{2}{c|}{
  \begin{tabular}{|*{4}{c|}}
\hline
    0.0 & 0.0 & 0.3 & 0.3  \\
    \hline
    0.0 & 0.7 & 0.7 & 0.3 \\
    \hline
    0.0 & 0.0 & 0.7 & 0.0  \\
    \hline
    0.0 & 0.0 & 0.0 & 0.0 \\
    \hline
\end{tabular}   }

&
    \multicolumn{2}{c|}{
\begin{tabular}{|*{4}{c|}}
\hline
    0.0 & 0.0 & 0.0 & 0.0  \\
    \hline
    0.0 & 1.0 & 1.0 & 0.0 \\
    \hline
    0.0 & 0.0 & 1.0 & 0.0  \\
    \hline
    0.0 & 0.0 & 0.0 & 0.0 \\
    \hline
\end{tabular}
}

&
    \multicolumn{2}{c}{
\begin{tabular}{|*{4}{c|}}
\hline
    0.0 & 0.0 & 0.0 & 0.0  \\
    \hline
    0.0 & 1.0 & 1.0 & 0.0 \\
    \hline
    0.0 & 0.0 & 1.0 & 0.0  \\
    \hline
    0.0 & 0.0 & 0.0 & 0.0 \\
    \hline
\end{tabular}
}
\\
    & \multicolumn{2}{c|}{} & \multicolumn{2}{c|}{} & \multicolumn{2}{c}{}

\\
%2nd row -- hit SE
Hit SE, $h$

&
\begin{tabular}{|*{3}{c|}}
\hline
    0.0 & 0.0 & 0.0  \\
\hline
    0.7 & \circled{0.7}  & 0.0\\
\hline
    0.0 & 0.7 & 0.0 \\
\hline
\end{tabular}

&

\begin{tabular}{|*{3}{@{\kern0.4em}c@{\kern0.4em}|}}
\hline
    $-\infty$ & $-\infty$ & $-\infty$  \\
\hline
    0.7 & \circled{0.7}  & $-\infty$\\
\hline
    $-\infty$ & 0.7 & $-\infty$ \\
\hline
\end{tabular}
    
&

\begin{tabular}{|*{3}{c|}}
\hline
    0.0 & 0.0 & 0.0  \\
\hline
    0.7 & \circled{0.7}  & 0.0\\
\hline
    0.0 & 0.7 & 0.0 \\
\hline
\end{tabular}

&

\begin{tabular}{|*{3}{@{\kern0.4em}c@{\kern0.4em}|}}
\hline
    $-\infty$ & $-\infty$ & $-\infty$  \\
\hline
    0.7 & \circled{0.7}  & $-\infty$\\
\hline
    $-\infty$ & 0.7 & $-\infty$ \\
\hline
\end{tabular}

&
\begin{tabular}{|@{\kern0.4em}c@{\kern0.4em}|c|@{\kern0.4em}c@{\kern0.4em}|}
\hline
    $-\infty$ & 0.1 & 0.1  \\
\hline
    0.7 & \circled{0.7}  & 0.1\\
\hline
    $-\infty$ & 0.7 & $-\infty$ \\
\hline
\end{tabular}
&
\begin{tabular}{|@{\kern0.4em}c@{\kern0.4em}|c|@{\kern0.4em}c@{\kern0.4em}|}
\hline
    $-\infty$ & 0.1 & 0.1  \\
\hline
    0.7 & \circled{0.7}  & 0.7\\
\hline
    $-\infty$ & 0.7 & $-\infty$ \\
\hline
\end{tabular}
\\
Miss SE, $m$
&
\begin{tabular}{|*{3}{c|}}
\hline
    0.0 & 0.7 & 0.7  \\
\hline
    0.0 & \circled{0.0}  & 0.7\\
\hline
    0.0 & 0.0 & 0.0 \\
\hline
\end{tabular}

&

\begin{tabular}{|*{3}{@{\kern0.4em}c@{\kern0.4em}|}}
\hline
    $-\infty$ & 0.7 & 0.7  \\
\hline
    $-\infty$ & \circled{$-\infty$}  & 0.7\\
\hline
    $-\infty$ & $-\infty$ & $-\infty$ \\
\hline
\end{tabular}
    
&

\begin{tabular}{|*{3}{c|}}
\hline
    0.0 & 0.7 & 0.7  \\
\hline
    0.0 & \circled{0.0}  & 0.7\\
\hline
    0.0 & 0.0 & 0.0 \\
\hline
\end{tabular}

&

\begin{tabular}{|*{3}{@{\kern0.4em}c@{\kern0.4em}|}}
\hline
    $-\infty$ & 0.7 & 0.7  \\
\hline
    $-\infty$ & \circled{$-\infty$}  & 0.7\\
\hline
    $-\infty$ & $-\infty$ & $-\infty$ \\
\hline
\end{tabular}

&
\begin{tabular}{|@{\kern0.4em}c@{\kern0.4em}|c|@{\kern0.4em}c@{\kern0.4em}|}
\hline
    $-\infty$ & 0.7 & 0.7  \\
\hline
    0.1 & \circled{0.1}  & 0.7\\
\hline
    $-\infty$ & 0.1 & $-\infty$ \\
\hline
\end{tabular}
&
\begin{tabular}{|@{\kern0.4em}c@{\kern0.4em}|c|@{\kern0.4em}c@{\kern0.4em}|}
\hline
    $-\infty$ & 0.7 & 0.7  \\
\hline
    0.1 & \circled{0.1}  & 0.7\\
\hline
    $-\infty$ & 0.1 & $-\infty$ \\
\hline
\end{tabular}
\\

Erosion of $f$ by $h$, $f\ominus h$

&

% \begin{tabular}{|@{\kern0.4em}c@{\kern0.4em}|@{\kern0.4em}c@{\kern0.4em}|}
% \hline
%     $-0.7$ & $0.0$  \\
%     \hline
%     $-0.7$ & $-0.7$\\
%     \hline
% \end{tabular}

\begin{tabular}{|c|@{\kern0.8em}c@{\kern0.8em}|@{\kern0.8em}c@{\kern0.8em}|c|}
\hline
* & * & * & * \\
\hline
   * &  -0.7 & 0.0 & *  \\
    \hline
  * &   -0.7 & -0.7 & *\\
    \hline
* & * & * & * \\
\hline
\end{tabular}

&

% \begin{tabular}{|@{\kern0.4em}c@{\kern0.4em}|@{\kern0.4em}c@{\kern0.4em}|}
% \hline
%     $-0.7$ & $0.0$  \\
%     \hline
%     $-0.7$ & $-0.7$\\
%     \hline
% \end{tabular}

\begin{tabular}{|c|@{\kern0.8em}c@{\kern0.8em}|@{\kern0.8em}c@{\kern0.8em}|c|}
\hline
* & * & * & * \\
\hline
   * &  -0.7 & 0.0 & *  \\
    \hline
  * &   -0.7 & -0.7 & *\\
    \hline
* & * & * & * \\
\hline
\end{tabular}

&
\begin{tabular}{|c|@{\kern0.8em}c@{\kern0.8em}|@{\kern0.8em}c@{\kern0.8em}|c|}
\hline
* & * & * & * \\
\hline
   * &  -0.7 & 0.0 & *  \\
    \hline
  * &   -0.7 & -0.7 & *\\
    \hline
* & * & * & * \\
\hline
\end{tabular}
&
\begin{tabular}{|c|@{\kern0.8em}c@{\kern0.8em}|@{\kern0.8em}c@{\kern0.8em}|c|}
\hline
* & * & * & * \\
\hline
   * &  -0.7 & 0.3 & *  \\
    \hline
  * &   -0.7 & -0.7 & *\\
    \hline
* & * & * & * \\
\hline
\end{tabular}
&
\begin{tabular}{|c|@{\kern0.8em}c@{\kern0.8em}|@{\kern0.8em}c@{\kern0.8em}|c|}
\hline
* & * & * & * \\
\hline
   * &  -0.7 & -0.1 & *  \\
    \hline
  * &   -0.7 & -0.7 & *\\
    \hline
* & * & * & * \\
\hline
\end{tabular}

&
\begin{tabular}{|c|@{\kern0.8em}c@{\kern0.8em}|@{\kern0.8em}c@{\kern0.8em}|c|}
\hline
* & * & * & * \\
\hline
   * &  -0.7 & -0.7 & *  \\
    \hline
  * &   -0.7 & -0.7 & *\\
    \hline
* & * & * & * \\
\hline
\end{tabular}
\\

Dilation of $f$ by $m$, $f\oplus m$

&

% \begin{tabular}{|c|c|}
% \hline
%     $1.4$ & $1.0$  \\
%     \hline
%     $1.4$ & $1.4$\\
%     \hline
% \end{tabular}

\begin{tabular}{|c|@{\kern0.8em}c@{\kern0.8em}|@{\kern0.8em}c@{\kern0.8em}|c|}
\hline
* & * & * & * \\
\hline
   * & 1.4 & 1.0 & *  \\
    \hline
  * &  1.4 & 1.4 & *\\
    \hline
* & * & * & * \\
\hline
\end{tabular}

&

\begin{tabular}{|c|@{\kern0.8em}c@{\kern0.8em}|@{\kern0.8em}c@{\kern0.8em}|c|}
\hline
* & * & * & * \\
\hline
   * & 1.4 & 1.0 & *  \\
    \hline
  * &  1.4 & 1.4 & *\\
    \hline
* & * & * & * \\
\hline
\end{tabular}

&
% \begin{tabular}{|c|c|}
% \hline
%     $1.7$ & $1.0$  \\
%     \hline
%     $1.7$ & $1.7$\\
%     \hline
% \end{tabular}

\begin{tabular}{|c|@{\kern0.8em}c@{\kern0.8em}|@{\kern0.8em}c@{\kern0.8em}|c|}
\hline
* & * & * & * \\
\hline
   * & 1.7 & 1.0 & *  \\
    \hline
  * &  1.7 & 1.7 & *\\
    \hline
* & * & * & * \\
\hline
\end{tabular}

&
% \begin{tabular}{|c|c|}
% \hline
%     $1.7$ & $0.7$  \\
%     \hline
%     $1.7$ & $1.7$\\
%     \hline
% \end{tabular}

\begin{tabular}{|c|@{\kern0.8em}c@{\kern0.8em}|@{\kern0.8em}c@{\kern0.8em}|c|}
\hline
* & * & * & * \\
\hline
   * & 1.7 & 0.7 & *  \\
    \hline
  * &  1.7 & 1.7 & *\\
    \hline
* & * & * & * \\
\hline
\end{tabular}

&
% \begin{tabular}{|c|c|}
% \hline
%     $1.7$ & $1.1$  \\
%     \hline
%     $1.7$ & $1.7$\\
%     \hline
% \end{tabular}

\begin{tabular}{|c|@{\kern0.8em}c@{\kern0.8em}|@{\kern0.8em}c@{\kern0.8em}|c|}
\hline
* & * & * & * \\
\hline
   * & 1.7 & 1.1 & *  \\
    \hline
  * &  1.7 & 1.7 & *\\
    \hline
* & * & * & * \\
\hline
\end{tabular}

&
% \begin{tabular}{|c|c|}
% \hline
%     $1.7$ & $1.1$  \\
%     \hline
%     $1.7$ & $1.7$\\
%     \hline
% \end{tabular}

\begin{tabular}{|c|@{\kern0.8em}c@{\kern0.8em}|@{\kern0.8em}c@{\kern0.8em}|c|}
\hline
* & * & * & * \\
\hline
   * & 1.7 & 1.1 & *  \\
    \hline
  * &  1.7 & 1.7 & *\\
    \hline
* & * & * & * \\
\hline
\end{tabular}

\\
hit-or-miss transform, $f\odot(h,m)$
&

% \begin{tabular}{|@{\kern0.4em}c@{\kern0.4em}|@{\kern0.4em}c@{\kern0.4em}|}
% \hline
%     $-2.1$ & $-1.0$  \\
%     \hline
%     $-2.1$ & $-2.1$\\
%     \hline
% \end{tabular}

\begin{tabular}{|c|@{\kern0.8em}c@{\kern0.8em}|@{\kern0.8em}c@{\kern0.8em}|c|}
\hline
* & * & * & * \\
\hline
   * & -2.1 & -1.0 & *  \\
    \hline
  * &  -2.1 & -2.1 & *\\
    \hline
* & * & * & * \\
\hline
\end{tabular}

&

\begin{tabular}{|c|@{\kern0.8em}c@{\kern0.8em}|@{\kern0.8em}c@{\kern0.8em}|c|}
\hline
* & * & * & * \\
\hline
   * & -2.1 & -1.0 & *  \\
    \hline
  * &  -2.1 & -2.1 & *\\
    \hline
* & * & * & * \\
\hline
\end{tabular}

&
% \begin{tabular}{|@{\kern0.4em}c@{\kern0.4em}|@{\kern0.4em}c@{\kern0.4em}|}
% \hline
%     $-2.4$ & $-1.0$  \\
%     \hline
%     $-2.4$ & $-2.4$\\
%     \hline
% \end{tabular}

\begin{tabular}{|c|@{\kern0.8em}c@{\kern0.8em}|@{\kern0.8em}c@{\kern0.8em}|c|}
\hline
* & * & * & * \\
\hline
   * & -2.4 & -1.0 & *  \\
    \hline
  * &  -2.4 & -2.4 & *\\
    \hline
* & * & * & * \\
\hline
\end{tabular}

&
% \begin{tabular}{|@{\kern0.4em}c@{\kern0.4em}|@{\kern0.4em}c@{\kern0.4em}|}
% \hline
%     $-2.4$ & $-0.4$  \\
%     \hline
%     $-2.4$ & $-2.4$\\
%     \hline
% \end{tabular}

\begin{tabular}{|c|@{\kern0.8em}c@{\kern0.8em}|@{\kern0.8em}c@{\kern0.8em}|c|}
\hline
* & * & * & * \\
\hline
   * & -2.4 & -0.4 & *  \\
    \hline
  * &  -2.4 & -2.4 & *\\
    \hline
* & * & * & * \\
\hline
\end{tabular}

&
% \begin{tabular}{|@{\kern0.4em}c@{\kern0.4em}|@{\kern0.4em}c@{\kern0.4em}|}
% \hline
%     $-2.4$ & $-1.2$  \\
%     \hline
%     $-2.4$ & $-2.4$\\
%     \hline
% \end{tabular}

\begin{tabular}{|c|@{\kern0.8em}c@{\kern0.8em}|@{\kern0.8em}c@{\kern0.8em}|c|}
\hline
* & * & * & * \\
\hline
   * & -2.4 & -1.2 & *  \\
    \hline
  * &  -2.4 & -2.4 & *\\
    \hline
* & * & * & * \\
\hline
\end{tabular}

&
% \begin{tabular}{|@{\kern0.4em}c@{\kern0.4em}|@{\kern0.4em}c@{\kern0.4em}|}
% \hline
%     $-2.4$ & $-1.8$  \\
%     \hline
%     $-2.4$ & $-2.4$\\
%     \hline
% \end{tabular}

\begin{tabular}{|c|@{\kern0.8em}c@{\kern0.8em}|@{\kern0.8em}c@{\kern0.8em}|c|}
\hline
* & * & * & * \\
\hline
   * & -2.4 & -1.8 & *  \\
    \hline
  * &  -2.4 & -2.4 & *\\
    \hline
* & * & * & * \\
\hline
\end{tabular}

\\

Comments 
&
\multicolumn{2}{p{5cm}|}{SEs with and without DNC Produces the same results when the SE including $0$s fits perfectly in the image.}

&
\multicolumn{2}{p{5cm}|}{SE with DNC considers only foreground and background elements and yields better results}

&
\multicolumn{2}{p{5cm}}{Intersecting SEs are unable to find a match for both foreground and background structures.}

    \end{tabular}%
    % }
\caption{Grayscale hit-or-miss transform illustrating the importance of DNC and non-intersecting condition with an example of top-right corner detection. In SEs, $-\infty$ is used for DNC and the origins are marked with a circle. The first two columns show the case when the SEs including its $0$s exactly fit in the image, $f$. SEs with and without DNC produce the same results as expected. Third and fourth columns are for the case where SEs for hit and miss fit below and above, respectively in the target area (top-right $3\times 3$ window) of the input image. Without DNC, $0$s (vs. $0.7$) in $h$ determine the output, which remains the same even though the input image is changed. On the other hand,  with DNC, the output latches on $0.7$s, not on $0$s in $h$ and varies with the change in input. Fifth and sixth columns compare the effect of non-intersecting and intersecting SEs. In the sixth column, erosion of $f$ by intersecting SE $h$ produces $-0.7$ for all cells, meaning no matching foreground-pattern found in any window of the input image. }
% While value in the top-right cell in the hit-miss transform is relatively lower, this result is misleading as the hit-or-miss transform need to match both foreground and background.

\label{fig:grayscaleMorphExample}
\end{figure*}

Similar to the binary case, the filters must be non-intersecting, i.e., satisfy the following constraints,
\[h(x,y)  \le m^c(x,y) \text{ or } m(x,y)  \le h^c(x,y)\]
where $h^c$ and $m^c$ are the complement of $h$ and $m$, respectively. This condition prevents the hit and miss SEs from contradicting each other. According to this condition, if $h(x,y) = 0.9$, then $m(x,y)$ must be less than $1-0.9$ or $0.1$ for an unit interval image.

\subsection{Properties of morphological operations}
Both grayscale erosion and dilation as well as the hit-or-miss transform are translation invariant, i.e.,
\[ (f+c) \ominus b = f\ominus b + c \text{ and }\] 
% and
\[(f+c) \oplus b = f\oplus b + c,\]
where c is an arbitrary value. Note that these operations as seen from Eqs.~(\ref{eq:grayscale-dilation}) and (\ref{eq:grayscale-erosion}) are not scale invariant. In contrast, convolution is scale invariant but not translation invariant.

\section{Methods}
\subsection{Morphological Shared Weight Neural Network}
Inspired by the success of shared weight CNNs on handwritten digit recognition (MNIST dataset) by LeCun in 1990 \cite{le1990constrained}, Gader et al. \cite{gader1994image} introduced morphology based image algebra network substituting convolution for morphological operations. Particularly, they used the hit-or-miss transform because of its ability to take into account both background and foreground of an object. This transform was extended with a power mean to soften the extremely sensitive max and min operations, where all the parameters including the exponents of the power mean were learned. In later works, Won et. al. \cite{won1995morphological} and Khabou et al. \cite{khabou1999morphological} used the standard hit-or-miss transform. None of these works considered the following aspects of hit-or-miss transform, non-intersecting condition and DNC. As illustrated with examples of binary and grayscale morphology in Figs.~\ref{fig:binaryMorphExample}~and~\ref{fig:grayscaleMorphExample}, 
DNC plays an important role in the design of an SE that helps disregard irrelevant parts of an image not necessary for finding a target pattern while keeping focus only on the relevant parts. Without a mechanism in place to provide for DNC, each element will be treated as a part of the target pattern and contribute to the output even if they are not. This can hurt the performance when there is a lot of variation in context and object shape and size, however they still might perform well for rigid pattern with fixed size and shape with little change in background and foreground. 

Figure~\ref{fig:grayscaleMorphExample} illustrates the role of non-intersecting condition and DNC in the grayscale hit-or-miss transform. Considering these conditions, 
we propose the following hit-or-miss transform
\begin{align}
    (f \odot (h,m))(x,y) = \min_{a,b \in D_{h_f}} (f(x+a,y+b) - h(a,b)) \nonumber\\
    -\max_{a,b \in D_{m_b}} (f(x+a,y+b) + m(a,b)),
    \label{eq:hitmissopt}
\end{align}
subject to 
\[ h(a,b) \ge m^c(a,b), \text{ or } m(a,b) \ge h^c(a,b),\]
where $x+a,y+b \in D_f$, $h^c(a,b)$ and $m^c(a,b)$ are complements of $h$ and $m$, and $h_f \subseteq h$ and $m_b \subseteq m$ are the foreground and background elements in $h$ and $m$, respectively. We remark that SEs learned without non-intersection condition may turn out to preserve this property, however it cannot be guaranteed so we make this condition explicit in our proposed definition. 
 
\subsection{Hit-or-Miss Transform Neuron}
A major challenge in enforcing the non-intersecting condition via complement according to Eq.~(\ref{eq:hitmissopt}) is computing the ranges for the image and SEs. This is because the ranges can be at different scales and they can vary across layers and from one iteration to the next due to updating of elements during optimization. To circumvent this issue, we take a more restrictive approach (analogous to binary morphology) where an element in a hit-or-miss transform is exclusively foreground, background, or DNC. We propose two algorithms, one with single SE incorporating only the non-intersecting condition and another with two SEs incorporating both the non-intersecting condition and DNC.

\subsubsection*{Single SE hit-or-miss transform} Let $f$ be an image and $w$ be an SE. The SE elements are partitioned into $w_h$ and $w_m$ such that their pairwise intersection is empty, where $w_h = \{w : w \le 0 \}$ and  $w_m = \{w : w \ge 0 \}$. The hit-or-miss neuron is defined as
\begin{align}
    (f \odot w)(x,y) & = \min_{a,b\in D_{w_h}} (f(x+a,y+b)+w_h(a,b)) \nonumber \\ & - \max_{c,d\in D_{w_m}}(f(x+c,y+d)+w_m(c,d)),
    \label{eq:hitmissSEv1}
\end{align}
\noindent where $w_h$ and $w_m$ conceptually correspond to foreground and background. This formulation has advantages: (i) implicit complementary conditions, (ii) fewer parameters, and (iii) fewer algebraic operations, thus less complexity and more computational efficiency. A caveat of this method is $0$ acts as a transition point between foreground and background so DNC cannot be enforced around this transition point, which otherwise would hinder switching of foreground elements to background and vice versa.

\subsubsection*{Dual SEs hit-or-miss transform}
Algorithm~\ref{alg:HMT-SEs} outlines the proposed algorithm. The algorithm takes an input image, the size of the SEs, and the threshold for DNC.
To enforce DNC and/or the non-intersecting condition, we take aid of two auxiliary variables, $a_h$ and $a_m$, initialized with zeros. The elements not part of $h$ and $m$ are assigned to $-\infty$ in $a_h$ and $a_m$, respectively. Then the hit is calculated as $\min(f - h - a_h)$ and miss as $\max(f + m + a_m)$. 

Because of separate SEs for foreground and background, an element can switch back and forth from one to another without transitioning through DNC region. However, once an element falls below the threshold and enters into the DNC non-optimization space, it cannot revert owing to the gradient being zero in this space.

\begin{algorithm}
\fontsize{10pt}{10pt}\selectfont
\caption{The hit-or-miss transform using two SEs}
\label{alg:HMT-SEs}
Input: Image $f$ and threshold for DNC, $th$.\\
Initialize two matrices, $h$ and $m$ (hit and miss SEs), pseudo-randomly w.r.t. a half-normal distribution.\\
% Subtract the the mean to center $h$ and $m$ at $0$.
% $h = h - \mu_h \text{ and } m = m - \mu_m$.\\
% Calculate the standard deviation of $h$ and $m$, $\sigma_h$ and $\sigma_m$\\
% Calculate the absolute value for threshold, $th = 0.5k(\sigma_h + \sigma_m)$ \\
Find the mask for DNC as the indices, $I_D$ of $D = \{x | x \ge \max(h, m) \text{ and } x \le th \}$.\\
Find the mask of non-foreground elements in $h$ as the indices, $I_{A_h}$ of $A_h = \{x | x \le \max(h,m) \text{ and } x \in h \}$\\
Initialize an auxiliary matrix $a_h$ of the same size as $h$ with $0$'s.\\
Set $a_h[I_D + I_{A_h}] = - \infty $\\
Calculate $ hit = \min (f - h - a_h)$\\
Find the mask of non-background elements in $m$ as the indices, $I_{A_m}$ of $A_m = \{x | x \le \max(h,m) \text{ and } x \in m \}$\\
Initialize an auxiliary matrix $a_m$ of the same size as $m$ with $0$'s.\\
Set $a_m[I_D + I_{A_m}] = -\infty $\\
Calculate hit $ miss = \max (f + m + a_m)$\\
Calculate the hit-or-miss transform as $f \odot (h,m) = hit + miss $, 
\end{algorithm}

\subsubsection*{Soft hit-or-miss (SHM)}
Eq.~(\ref{eq:hitmissopt}) for the hit-or-miss transform involves max and min operations, which are highly restrictive and overly sensitive to fluctuations and noise in the input. As for instance, a sudden fluctuation in just one pixel can change the output from a target shape being present to absent. To date, numerous variants and extensions have been put forth to ameliorate this issue. For example, Gader et al. \cite{gader1994image} generalized the hit-or-miss transform substituting max and min for the weighted power mean ($\frac{1}{n} \sum_{i=1}^n w_i x_i^p)^{\frac{1}{p}}$ that has limitations such as being undefined for $0$s and yielding complex output for negative inputs in the case of fractional $p$. Fuzzy and rank hit-or-miss transforms introduced in literature are not suitable for gradient based optimization \cite{perret2009robust}. 
% Therefore, we propose an extension of the hit-or-miss transform, referred to herein as \textit{soft hit-or-miss} (SHM), using a parametric soft-max and soft-min in place for max and min, respectively.
% \begin{align}
%     (f & \odot^s (h,m))(x,y) = \nonumber\\ & {\text{softmin}}_{a,b \in D_{h_f}} (f(x+a,y+b) - h(a,b)) \nonumber\\&
%     -{\text{softmax}}_{a,b \in D_{m_b}} (f(x+a,y+b) + m(a,b)),
%     \label{eq:shm}
% \end{align}
% subject to 
% \[ h(a,b) \ge m^c(a,b), \text{ or } m(a,b) \ge h^c(a,b).\]

Motivated by \cite{gader1994image}, we extend the hit-or-miss transform in Eq.~\ref{eq:hitmissopt}, referred to herein as \textit{soft hit-or-miss} (SHM), using a parametric soft-max and soft-min in place for max and min, respectively.
\begin{align}
    (f & \odot^s (h,m))(x,y) = \nonumber\\ & {\text{softmin}}_{a,b \in D_{h_f}} (f(x+a,y+b) - h(a,b)) \nonumber\\&
    -{\text{softmax}}_{a,b \in D_{m_b}} (f(x+a,y+b) + m(a,b)),
    \label{eq:shm}
\end{align}
subject to 
\[ h(a,b) \ge m^c(a,b), \text{ or } m(a,b) \ge h^c(a,b).\]

While there exists several formulae to define soft-min and soft-max, herein we opt for a generalized mean based on smooth-max function parameterized by $\alpha$ that is favorable to gradient based optimization,
\begin{equation}
    s_\alpha(x) = \frac{\sum x e^{\alpha x}}{\sum e^\alpha x},
    \label{eq:generalizedSmoothMax}
\end{equation}
where $\alpha \in \mathbb{R}$. 
This has an advantage over other generalized mean equations such as power and Lehmer means that it produces real valued output for non-negative valued inputs when fractional exponent is used whereas power and Lehmer means produce complex-valued results. Based on Eq.~(\ref{eq:generalizedSmoothMax}), the softmax and softmin operators are defined as
\[\text{softmax} = s_{smax,\alpha}(x) = \{s_\alpha(x) | \alpha \ge 0\},\]
\[\text{softmin} = s_{smin, \alpha}(x) = \{s_{-\alpha}(x) | \alpha \ge 0\}.\]

\subsection{Hit--or-miss transform inspired generalized convolution}
% This is not a new concept,
Previous works like Maragos have demonstrated equivalency between the binary hit-or-miss transform and the thresholded correlation (the linear correlation operation followed by thresholding) between a binary image and the binary hit-or-miss transform. \cite{590053}.
However, no direct equivalency can be established between real-valued convolution (linear operation) and grayscale hit-or-miss transform (non-linear operation). Herein, we aim to draw an analogy between these two operations w.r.t. the detection mechanism. While both consider foreground and background of the target structure,
% In this section, we show that like the hit-or-miss transform, convolution also considers both foreground and background of a target structure. 
they differ in that an SE in the hit-or-miss encodes the absolute level of target shape whereas a filter in convolution encodes relative importance/level. As a result, the hit-or-miss transform can provide an absolute measure of how the target shape fits in an image whereas convolution provides relative measure of correlation or the degree of matching. The hit-or-miss transform tells us whether an image fits a target pattern and the minimum offset between target shape and image. For example, perfect alignment will produce a value of $0$ in the hit and $-1$ in the miss for an image with input in an unit interval (as illustrated with an example in Fig.~\ref{fig:grayscaleMorphExample}). On the other hand, convolution will produce higher output for an image with target than non-target, so just looking at the convolution output, we cannot say whether a target shape is present in the image or not.

Given an image $f$, the convolution operation on this image w.r.t. a filter $w$ is
\[ (f * w) (x,y) = \sum_{a,b\in D_w} f(x-a,y-b) \ w(a,b)\]
This equation can be decomposed into two parts with positive and negative weights, respectively.
\begin{align}
     (f * w) (x,y)= \sum_{(a,b)\in D_{w_h}} f(x-a,y-b) \ w_h(a,b) \nonumber\\ - \sum_{(c,d)\in D_{w_m}} (-1) f(x-c,y-d) w_m(c,d),
    \label{eq:conv}
\end{align}
where $w_h = \{w : w > 0 \}$ and $w_m = \{w : w < 0 \}$. It is worth noting the structural similarity of hit and miss terms with those in the hit-or-miss transform. Since $f * w$ increases with increasing coefficient of $w_h$ and decreasing coefficient of $w_m$, $w_h$ and $w_m$ indicate the weight or the relative importance of the foreground and background elements of the target pattern, respectively. As such, non-negative weights are hit (foreground), non-positive weights are miss (background), and zeros act as DNC in convolution. Following CNN convention, we do not flip image nor filter in our implementation.

The linear operation, sum, in Eq.~\ref{eq:conv}, gives equal importance to all operands regardless of their values. 
Instead, we can use soft-min for the foreground in the first term so that those with smallest values dominate the results (akin to erosion). Similarly, the sum for the background/miss can be generalized with a soft-max operation so that those with largest values in the local neighborhood dominate the results (akin to dilation). We refer to this extension as \textit{generalized convolution 1} (GC1), denoted $*^{g1}$.
\begin{align}
     (f  *^{g1} w)(x,y) &  = n \left(s_{smin,\alpha_1} (f(x-a,y-b) w_h(a,b)) \nonumber \right. \\  & \left. - s_{smax,\alpha_2}((-1)f(x-c,y-d) w_m(c,d))\right),
\label{eq:gc1}
\end{align}
or
\begin{align}
     (f *^{g1}w) (x,y) & = n \left(s_{smin,\alpha_1} (f(x-a,y-b) w_h(a,b)) \right. \nonumber \\ & \left. + s_{smax,\alpha_2}(f(x-c,y-d) w_m(c,d))\right)
     \label{eq:gc1-alt}
\end{align}
\noindent where $s_{smin,\alpha_1}$ and $s_{smax,\alpha_2}$ are the softmax and softmin aggregation operations spanning between mean and max and between min and mean, respectively. Note that we apply a multiplication factor $n$ in Eqs.~(\ref{eq:gc1}) and (\ref{eq:gc1-alt}) so that it becomes convolution when $\alpha=0$. Eq.~(\ref{eq:gc1-alt}) has the computational advantage over Eq.~(\ref{eq:gc1}) as it requires computation only of the soft-min whereas Eq.~\ref{eq:gc1} involves both soft-max and soft-min. 

We propose an alternative definition of the GC that instead of decomposing the convolution operation analogous to the hit-or-miss transform, applies soft-max and soft-min directly to the standard convolution and then takes their sum,
\begin{align}
     (f  *^{g2} w) (x,y) & = n (s_{smin,\alpha_1} (f(x-a, y-b) w(a,b)) \nonumber \\ & + s_{smax,\alpha_2}( f(x-a,y-b) w(a,b))).
     \label{eq:GC2}
\end{align}
We refer to this operation as \textit{GC 2} (GC2). Next, we discuss how this extension will affect the gradient-descent based optimization, more specifically initialization. 

\subsubsection*{Optimization}
Recent advancements and key insights into the optimization process of a neural network such as initialization, skip connection in a residual network, and batch normalization contributed to achieving high performance. Kaiming He et al. \cite{he2015delving} showed that initializing weights such that the variance of the output of a layer remains the same as the input helps to keep the distribution of gradients unvaried across all layers. This addresses vanishing gradients, enabling training of deep neural networks. However, their analysis was limited to convolution with ReLu activation function. 

Convolution involves sum and product, both of which are linear operations and have closed form equations for variances (e.g., sum of variances for sum and product of variances for product). In contrast, there is no similar closed-form equation for max/min and generalized mean. Therefore, we model the variance in the form of $\sigma_{s,\alpha}^2 = an^b \sigma_x^2$ for different values of $\alpha$, where $n$ is the number of elements in a SE, and $a$ and $b$ are learned.  We used a synthetic dataset where $x$ is generated pseudo-randomly from a Gaussian distribution with unit variance and  $n = [3 \ 6 \ 9 \dots 24]^2$. Table \ref{tab:generalized-mean-distr} lists the ratio of the input and output variances of Eq.~\ref{eq:generalizedSmoothMax} for different $\alpha$.

\begin{table}[htbp]
  \centering
  \renewcommand{\arraystretch}{1.6}
  \caption{Variance of the smooth-max function, $s_\alpha$ vs. $\alpha$}
    \begin{tabular}{l *{5}{c}}
    \toprule
    $\alpha$ &  $0$ & 
    % $\pm 0.25$ &
    $\pm 0.5$ &  
    % $\pm 0.75$ & 
    $\pm 1$ & $\pm 2$ & 
    % $\pm 4$ & $\pm 8$ & 
    $\pm \infty$ (max/min)\\
    \midrule
    $\sigma_{s_\alpha}'\left( = \sigma_{s_\alpha}^2 / \sigma_x^2\right)$ & $\frac{1}{n}$ & 
    % $\frac{1.0945}{n^{0.9933}}$ & 
    $\frac{1.32}{n^{0.95}}$ & 
    % $\frac{1.4712}{n^{0.8698}}$ & 
    $\frac{1.44}{n^{0.74}}$ & $\frac{0.82}{n^{0.32}}$ & 
    % $\frac{0.5585 }{n^{0.1695}}$ & $\frac{0.5789}{n^{0.2164}}$ & 
    $\frac{0.60}{n^{0.24}}$ \\
    % $\frac{0.5844}{n^{0.2392}}$
    % complementary variance =  0.6826
    % $\sigma_{s_\alpha}^2 / \sigma_x^2$ & $\frac{1}{n}$ &  & $\frac{1.0662}{n^{0.8630}}$  &  &  &  &  &  &  $\frac{0.8896}{n^{0.4071}}$\\    
    \bottomrule
    \end{tabular}%
  \label{tab:generalized-mean-distr}%
\end{table}%

Another challenge with finding exact criteria for initialization is the interdependency of terms. As we know, the variance of $z = x \pm y$ is  $\sigma_z^2 = \sigma_x^2 + \sigma_y^2 \pm 2\sigma_{xy}^2,$ where $\sigma_{xy}^2$ is the covariance between $x$ and $y$. When $x$ and $y$ are independent, their covariance will be zero, and the variance of $z$ can be obtained directly by summing up the variance of individual components. However, this is not the case for hit-or-miss transform (e.g., Eq.~(\ref{eq:hitmissopt})) and extensions (e.g., Eq.~(\ref{eq:gc1-alt})), where $f$ exists in both hit and miss terms. Since $\alpha$ changes the distribution, which in turn changes the covariance,  the analysis is very complicated. Herein, we simplify variance analysis by ignoring the covariance term and using modeled equations for the generalized mean. The Appendix provides initialization criteria for extensions of hit-or-miss transforms and convolution, which also apply to standard operations. 

\begin{threeparttable}
\scriptsize
\setlength\tabcolsep{0pt} % let LaTeX determine whitespace between columns
  \centering
  \renewcommand{\arraystretch}{1.3}
  \caption{Mini-VGG (4 layer NN)}
    \begin{tabular}{cc}
    \toprule
    Layer & Filter size \\
    % Corresponding element in the target pattern
    \midrule
    \multirow{2}{*}{Input layer} & $28\times28\times1$ (MNIST/Fashion-MNIST)\\
    & $32 \times 32 \times 3$ (Cifar-10) \\
    HMC\tnote{1} \ layer + BN\tnote{2} + ReLU\tnote{3} & $3 \times 3 \times 32$, padding:1  \\
    HMC\tnote{1}  \ layer + BN + ReLU & $3 \times 3 \times 32$, padding=1  \\
    MaxPool & $2\times 2$ \\
    Dropout & 25\%  \\
    HMC\tnote{1} \ layer + BN + RelU & $3 \times 3 \times 64$, padding=1  \\
    HMC\tnote{1} \ layer + BN + ReLU & $3 \times 3 \times 64$, padding=1  \\
    MaxPool & $2\times 2$ \\
    Dropout & 25\%  \\
    Fully Connected Layer + BN + ReLU & 512  \\
    Dropout & 50\%  \\
    Softmax layer & 10\\
    \bottomrule
    \end{tabular}%
    \begin{tablenotes}
\item[1] HMC denotes the basic operation specific to a particular network, e.g., convolution in CNN and hit-or-miss in morphological NN;
\item[2] Batch-Normalization;
\item[3] Rectified Linear Unit.
\end{tablenotes}
  \label{tab:four-layers}%
% \end{table}%
\end{threeparttable}

\section{Experiments}\label{sec:exp}

In order to compare our proposed algorithms with its standard counterparts, for sake of an apples-to-apples comparison in a controlled fashion where we can responsibly account for all the moving parts, we consider both synthetic and real datasets. The synthetic dataset consists of a simple classification task with two fixed-shape objects such that all the methods can correctly classify the objects using a single layer, thus allowing us to visualize and interpret the learned SEs to shed light onto the inner workings of these algorithms.

We also evaluate the performance of our proposed algorithms in terms of classification accuracy on two benchmark datasets with varying context, shape, and size---from approximately fixed-sized, and rigid shaped objects with constant background in the Fashion-MNIST to complex background, variable size, and shaped objects in Cifar-10. These datasets and CNNs were carefully selected to enable the fairest comparison possible. The goal is to understand the benefits and drawbacks of morphology relative to convolution. In future work, we will focus more on obtaining state-of-the-art global neural morphological architectures such as GoogLeNet, ResNet, NASNets, and similar. The focus here is the fundamental value of morphology and how making it scale to deep contexts.

Since our focus is to compare different feature learning operations rather than other aspects of deep learning such as architecture or optimization algorithms, we  select a small VGG-like \cite{simonyan2014very} architecture with $4$ layers, referred to as mini-VGG (see Table~\ref{tab:four-layers} for its architecture). This small NN also allows us to have the same setup (e.g., hyper-parameters and optimization algorithm) for all experiments, including convolution and standard hit-or-miss. First, we provide an analysis of different initialization strategies followed by experiments on hit-or-miss transform and convolution and their extensions.

\subsection{Synthetic dataset}
\begin{figure*}
    \centering
    \scriptsize
\begin{tabular}{P{1.7cm} P{2.4cm} P{1.7cm} *{5}{P{1.7cm}}}
\multirow{2}{*}{
\includegraphics[width=0.1\textwidth]{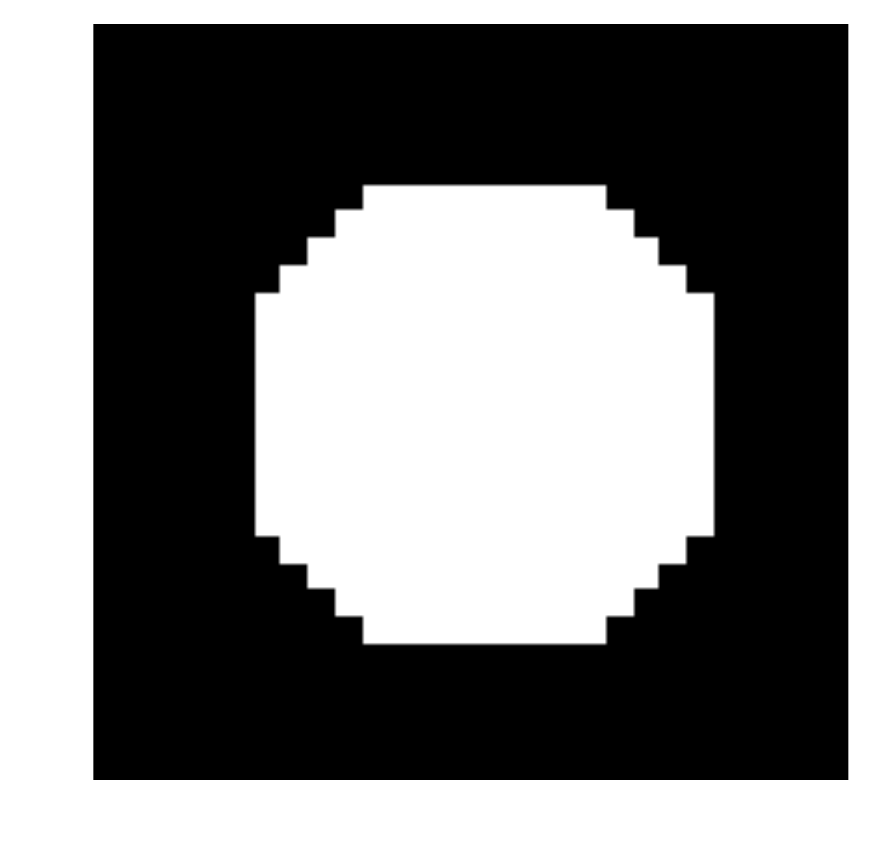}
}
&
\copyrightbox[l]{
\includegraphics[width=0.1\textwidth]{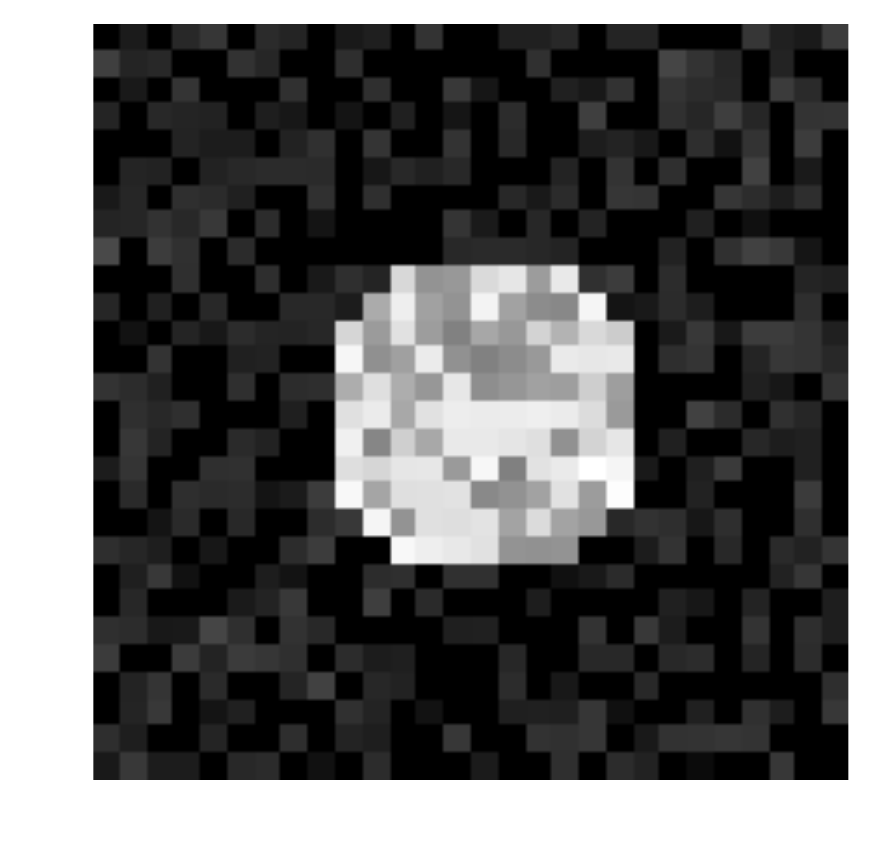}
}{Hit SE}
&
\includegraphics[width=0.1\textwidth]{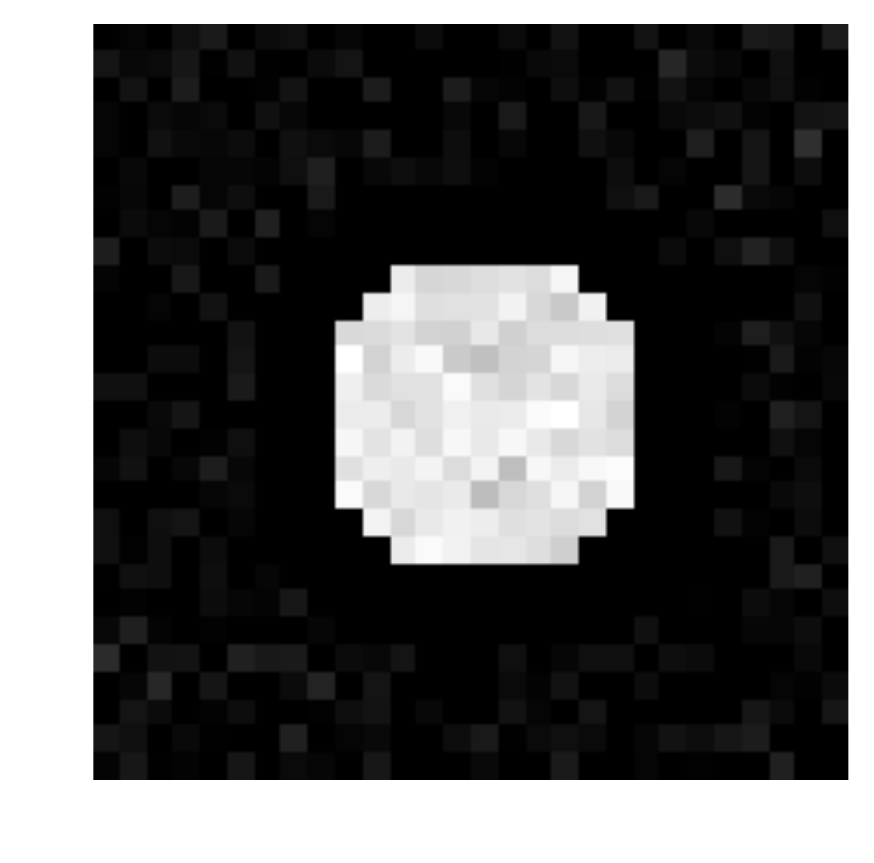}
&
\includegraphics[width=0.1\textwidth]{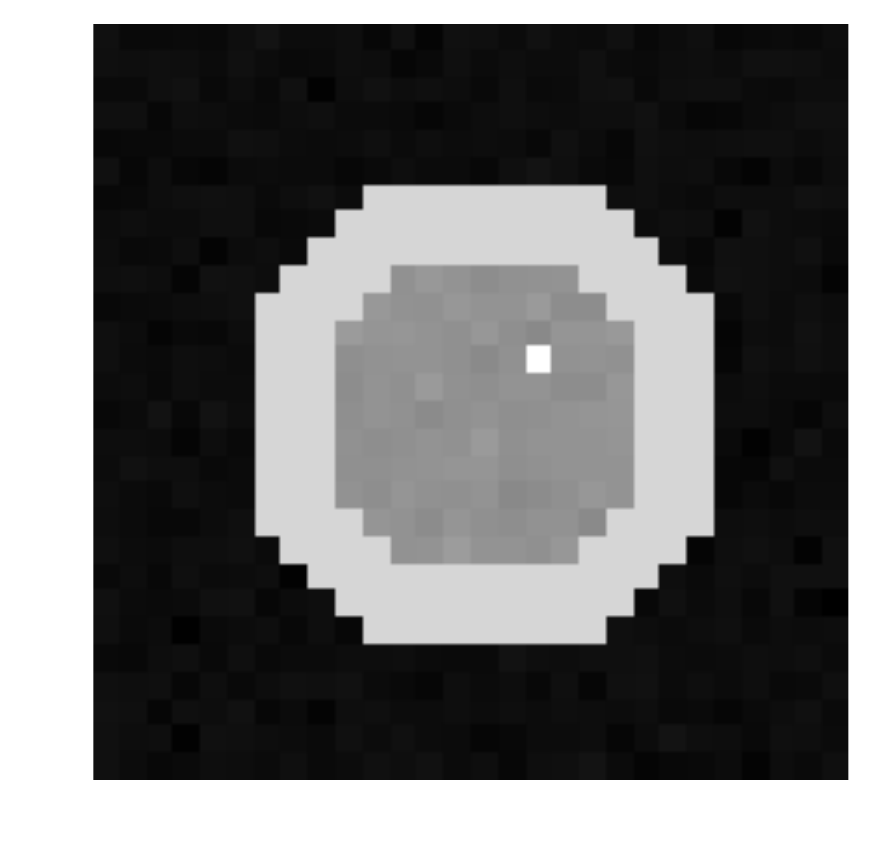}
& 
\includegraphics[width=0.1\textwidth]{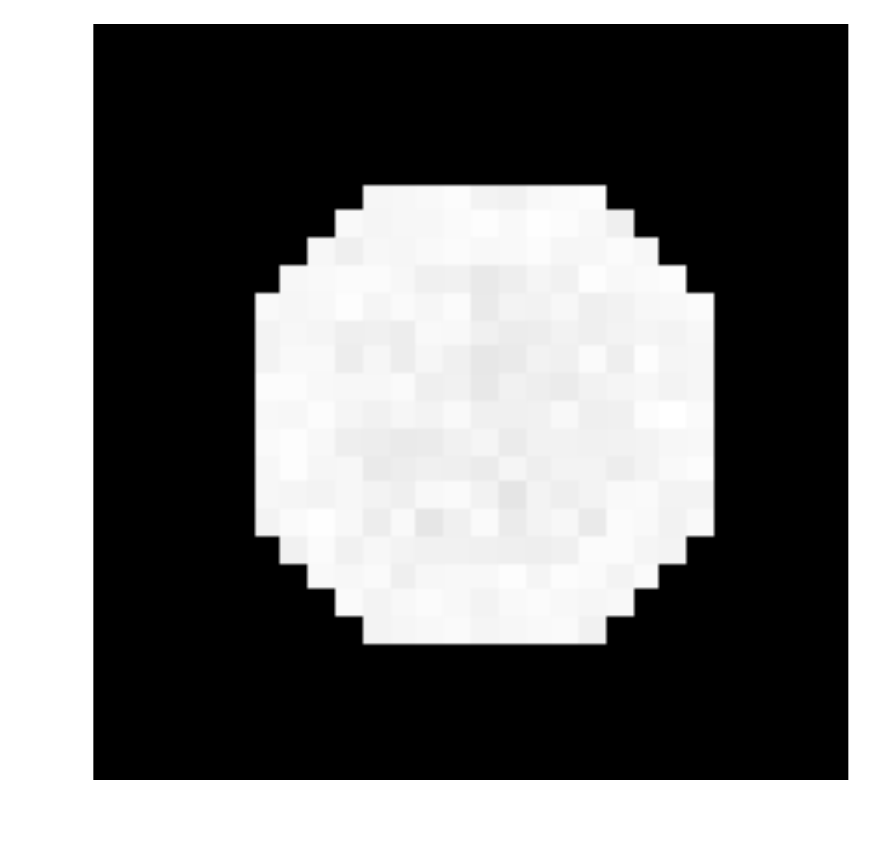}
&
\includegraphics[width=0.1\textwidth]{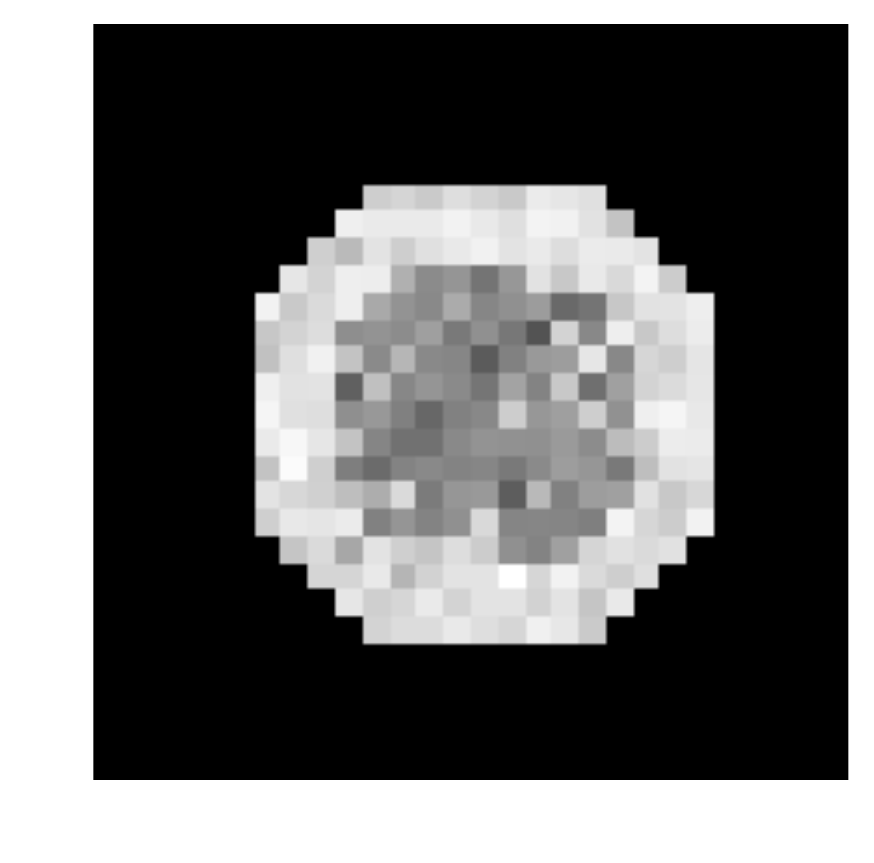}
&
\includegraphics[width=0.1\textwidth]{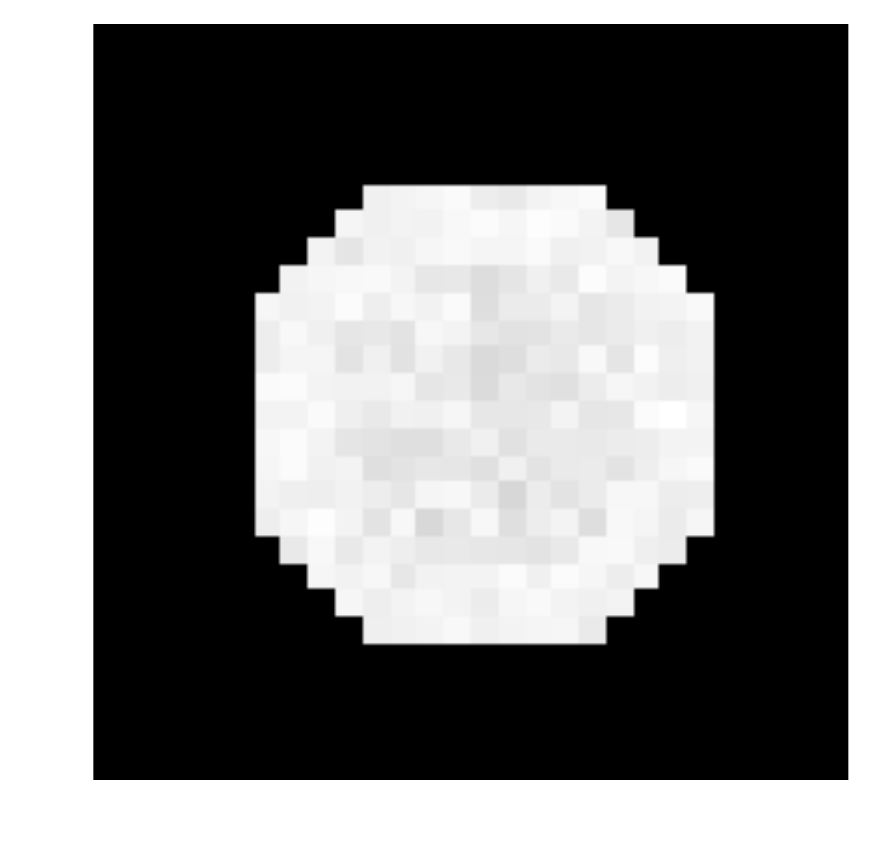}
&
\includegraphics[width=0.1\textwidth]{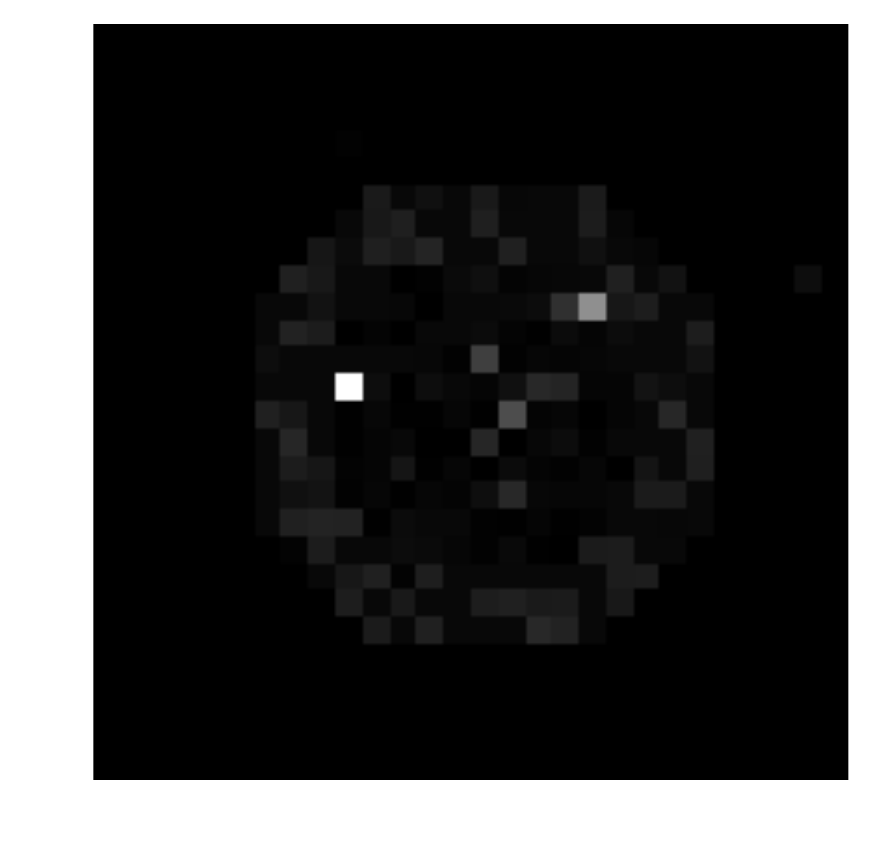}
% &
% \includegraphics[width=0.1\textwidth]{figures/hitmissSofthm-hit_1.pdf}
 \\
 & 
 \copyrightbox[l]{
\includegraphics[width=0.1\textwidth]{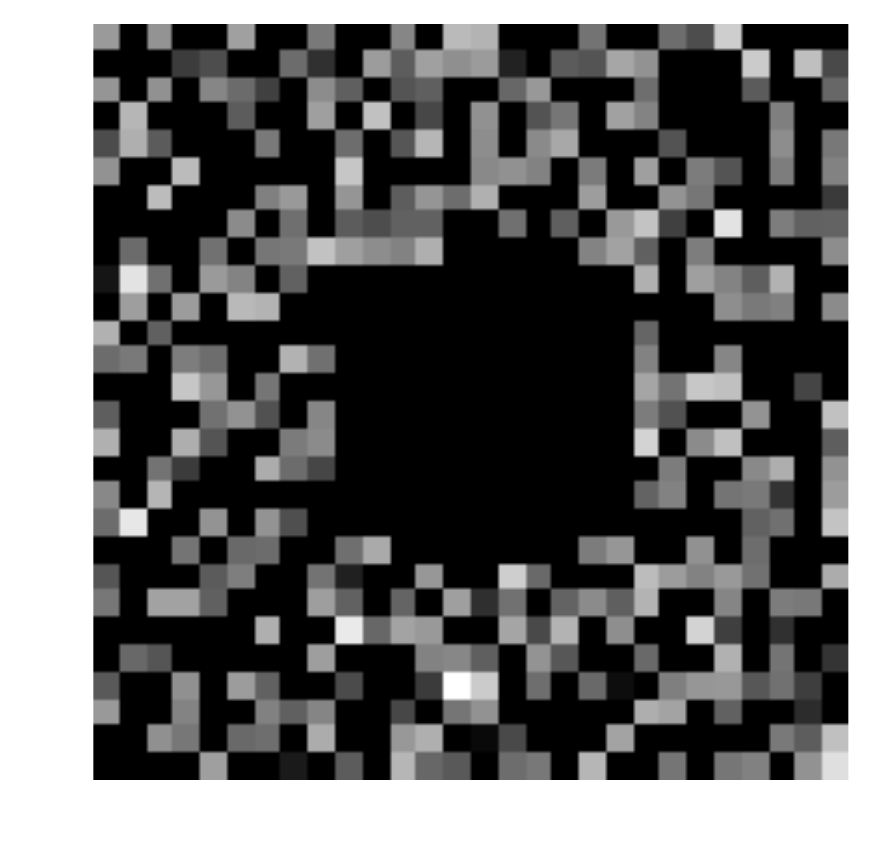}
}{Miss SE}
&
\includegraphics[width=0.1\textwidth]{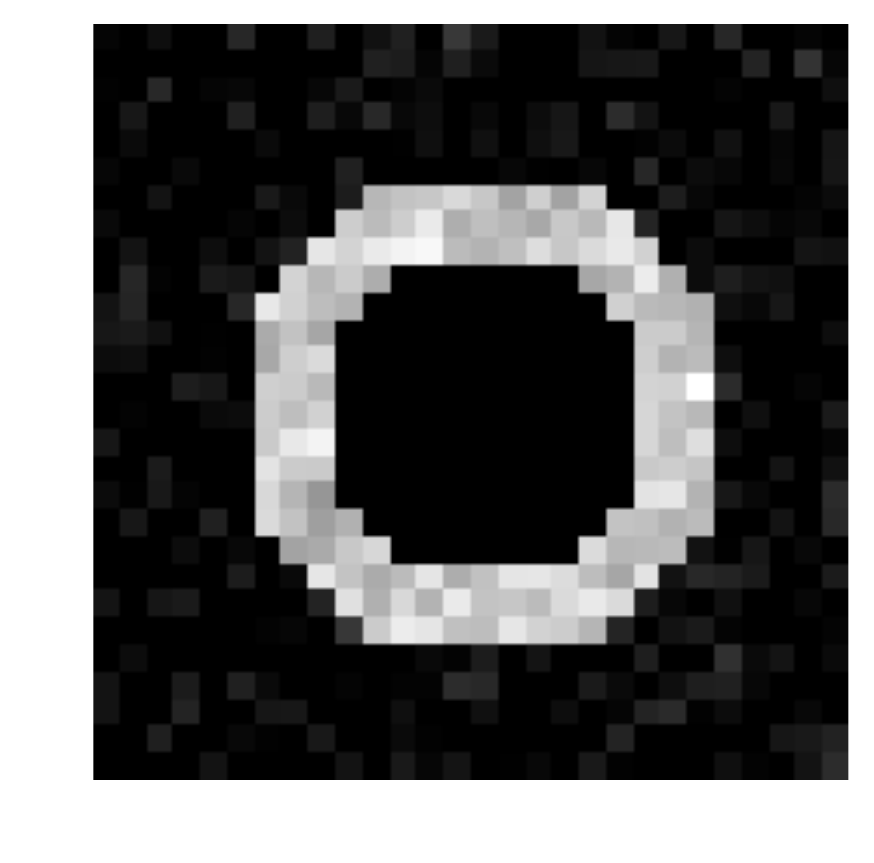}
&
\includegraphics[width=0.1\textwidth]{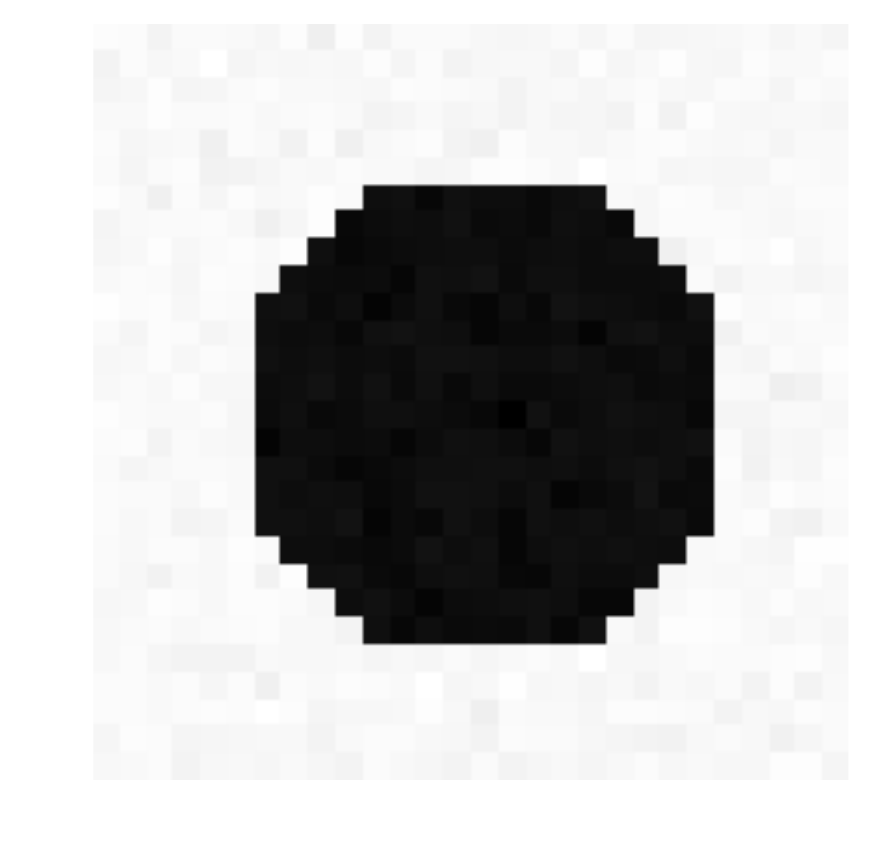}
& 
\includegraphics[width=0.1\textwidth]{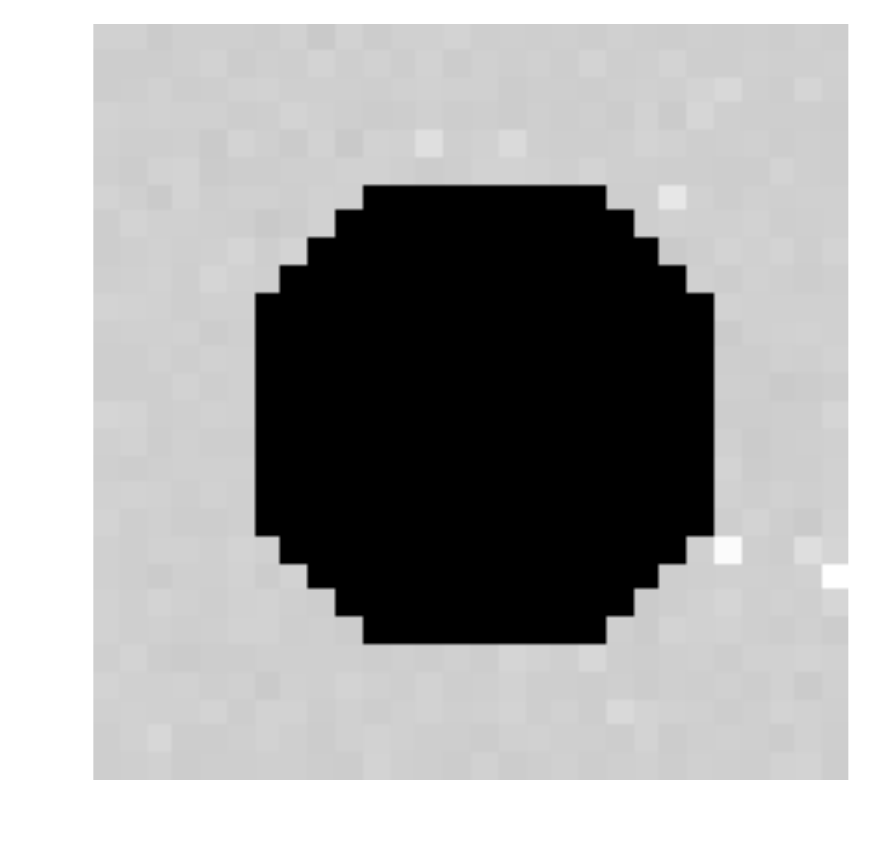}
&
\includegraphics[width=0.1\textwidth]{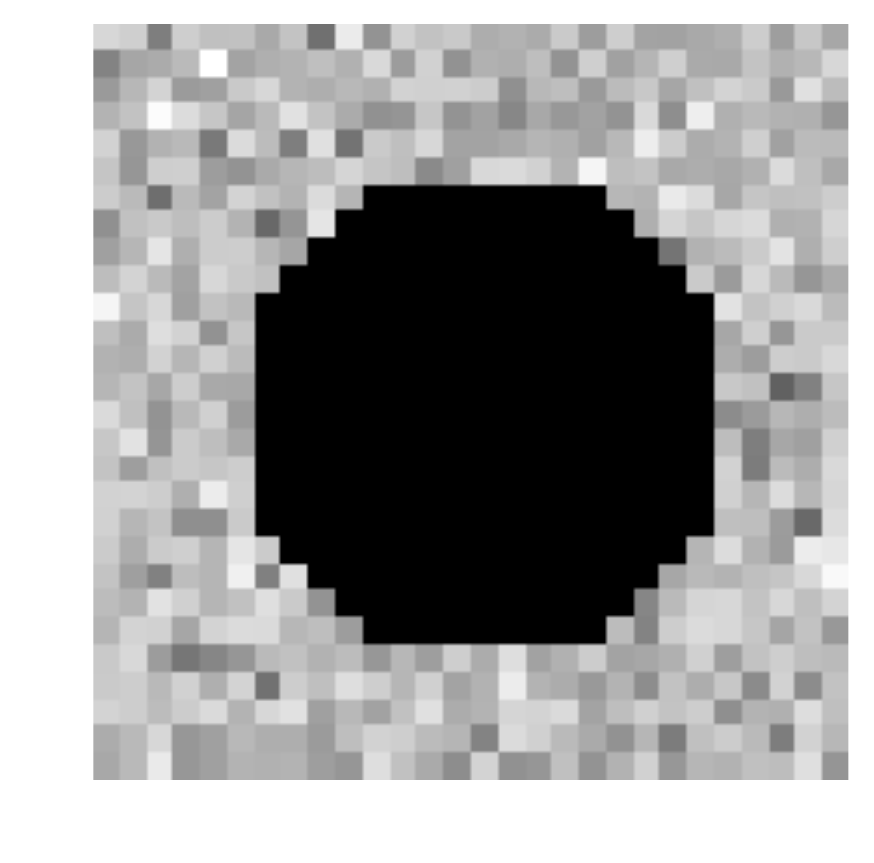}
&
\includegraphics[width=0.1\textwidth]{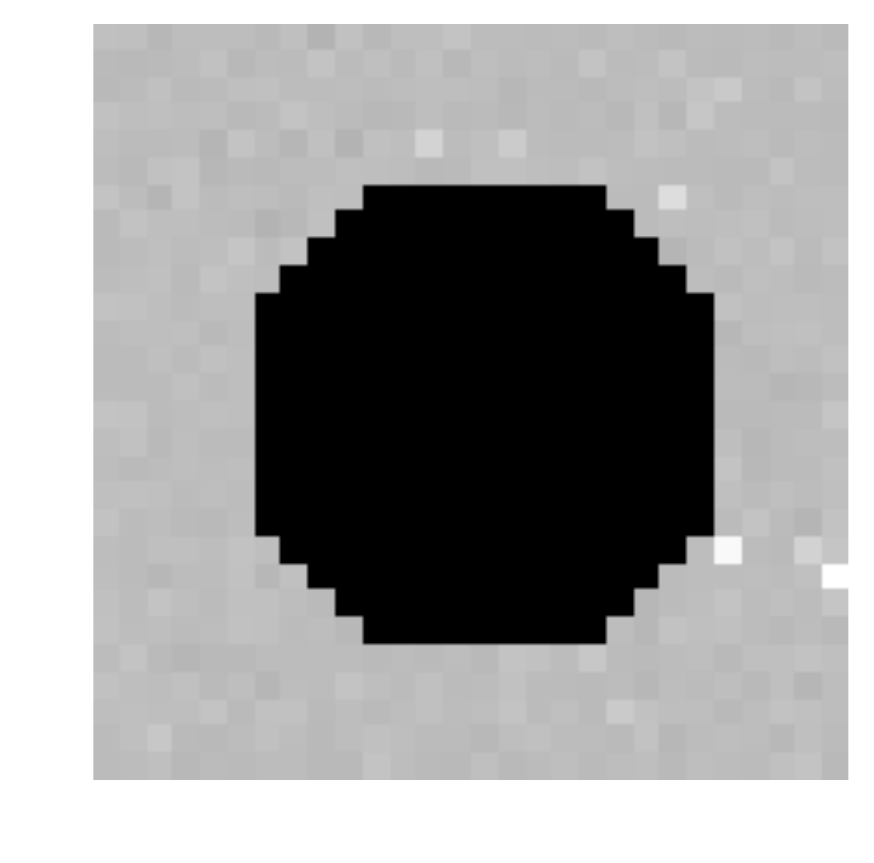}
&
\includegraphics[width=0.1\textwidth]{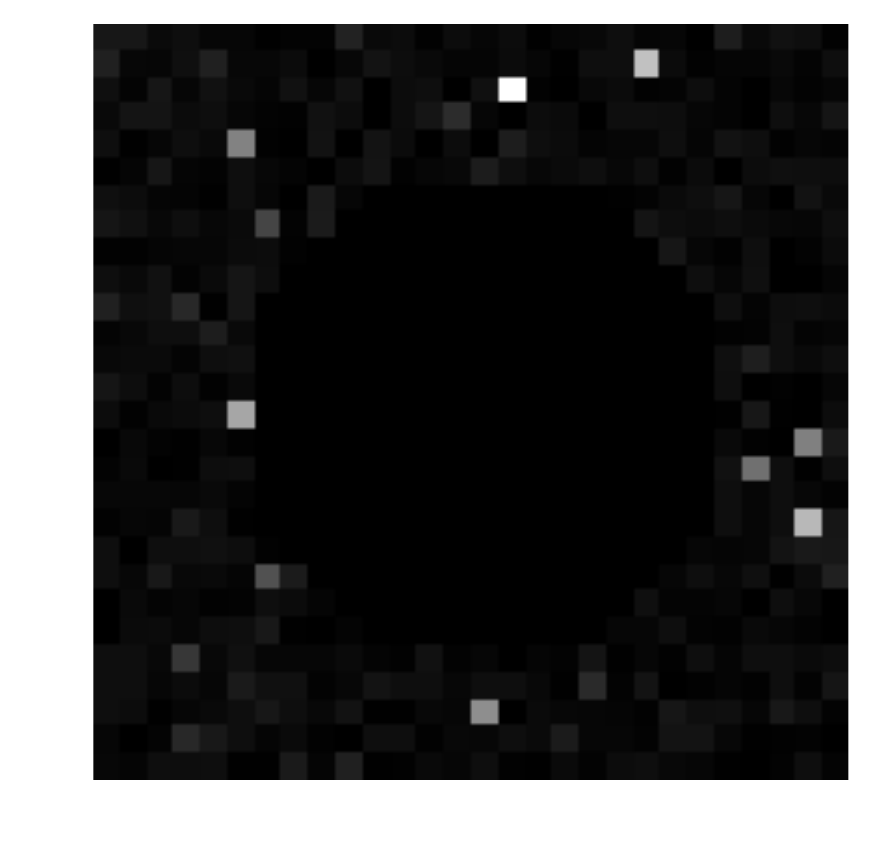}
% &
% \includegraphics[width=0.1\textwidth]{figures/hitmissSofthm-miss_1.pdf}
\\
\multirow{2}{*}{
\includegraphics[width=0.1\textwidth]{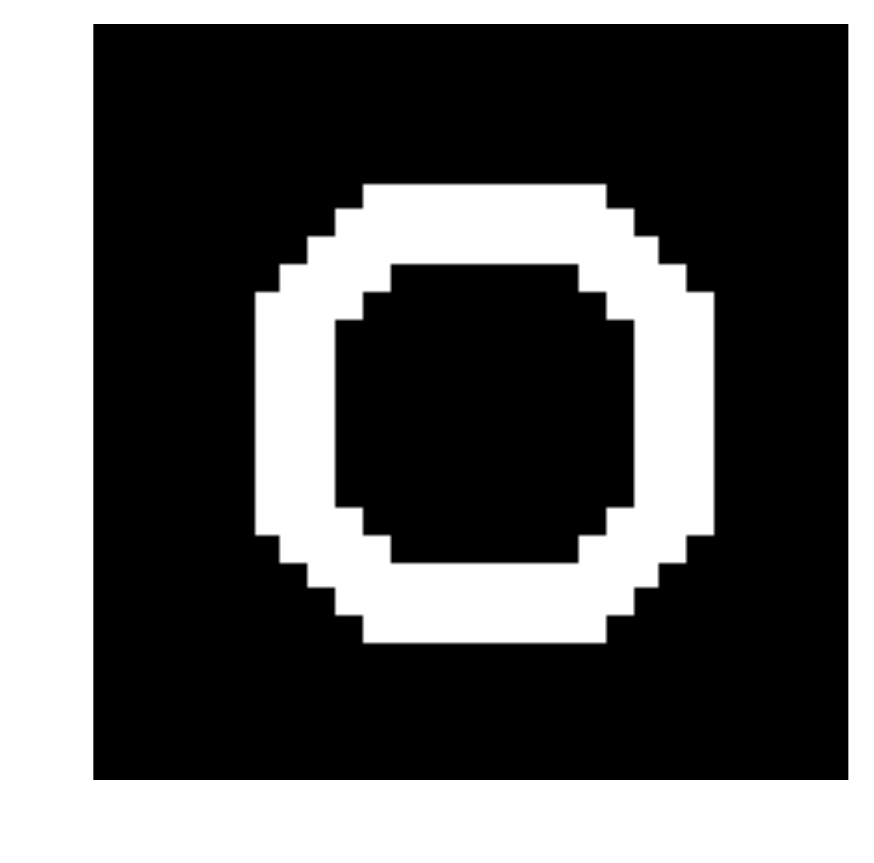}
}
&
\copyrightbox[l]{
\includegraphics[width=0.1\textwidth]{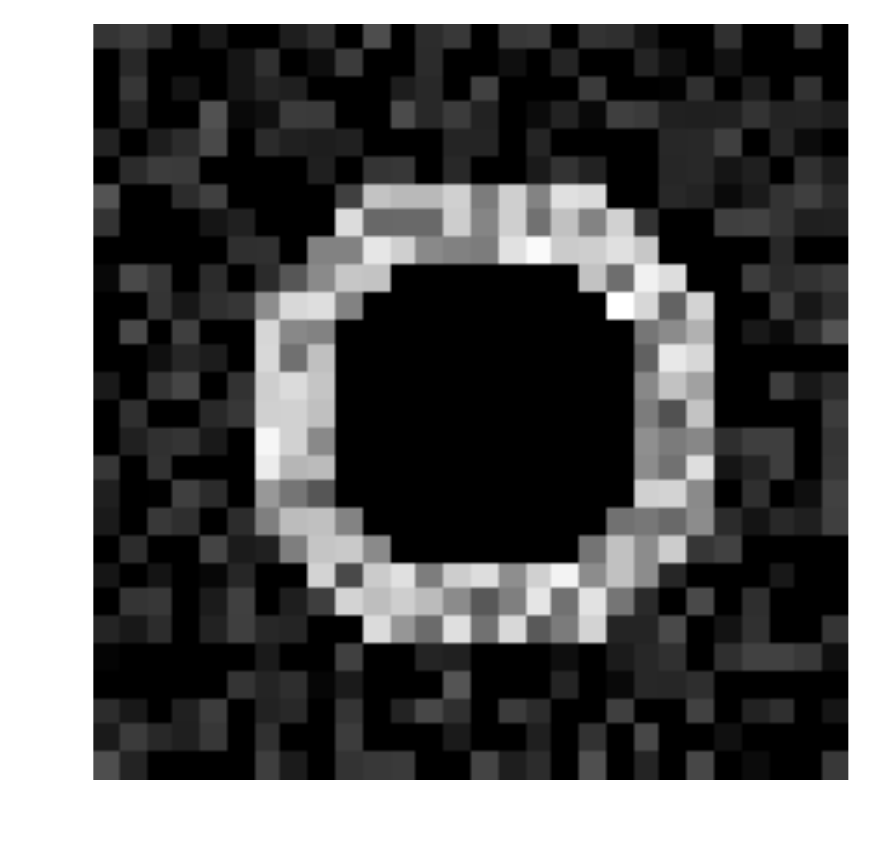}
}{Hit SE}
&
\includegraphics[width=0.1\textwidth]{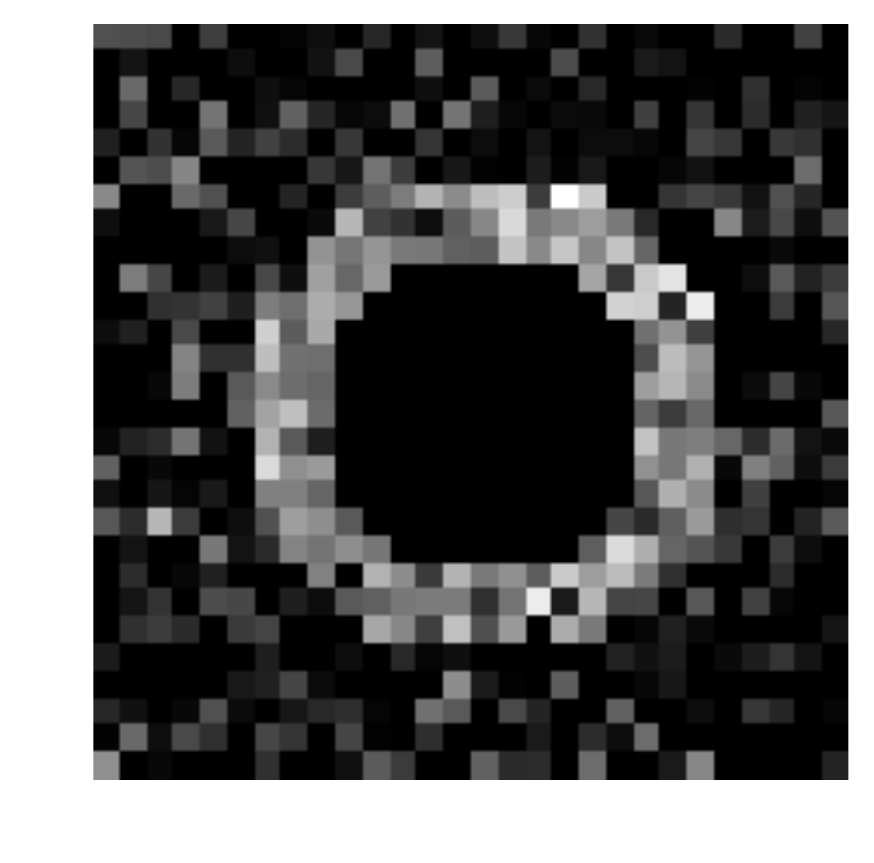}
&
\includegraphics[width=0.1\textwidth]{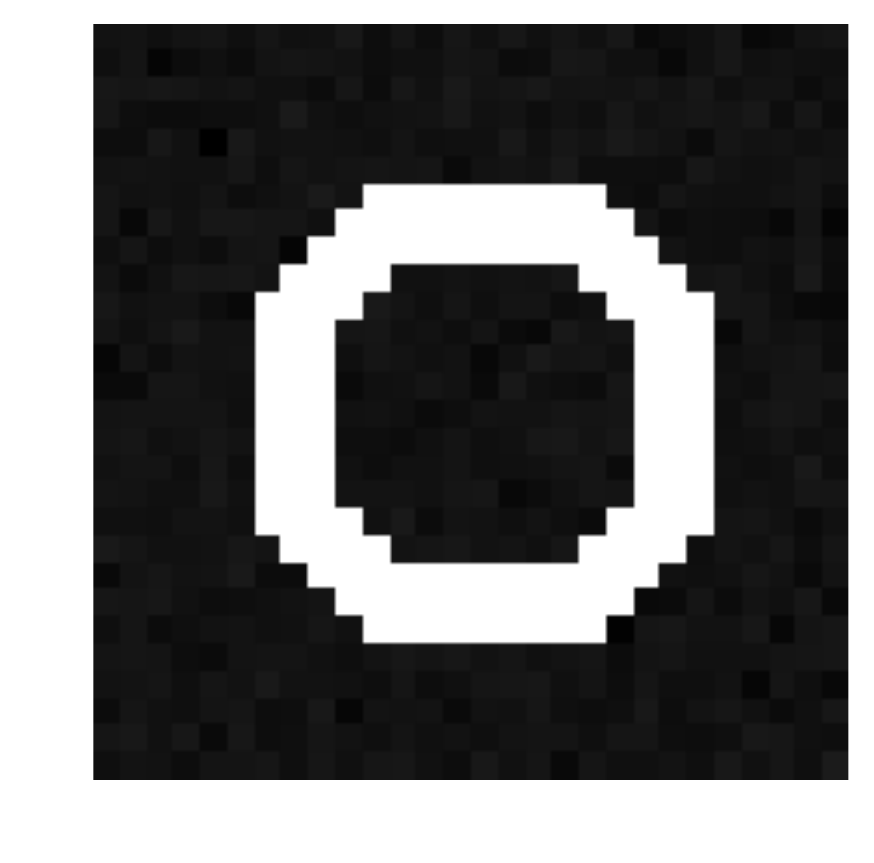}
& 
\includegraphics[width=0.1\textwidth]{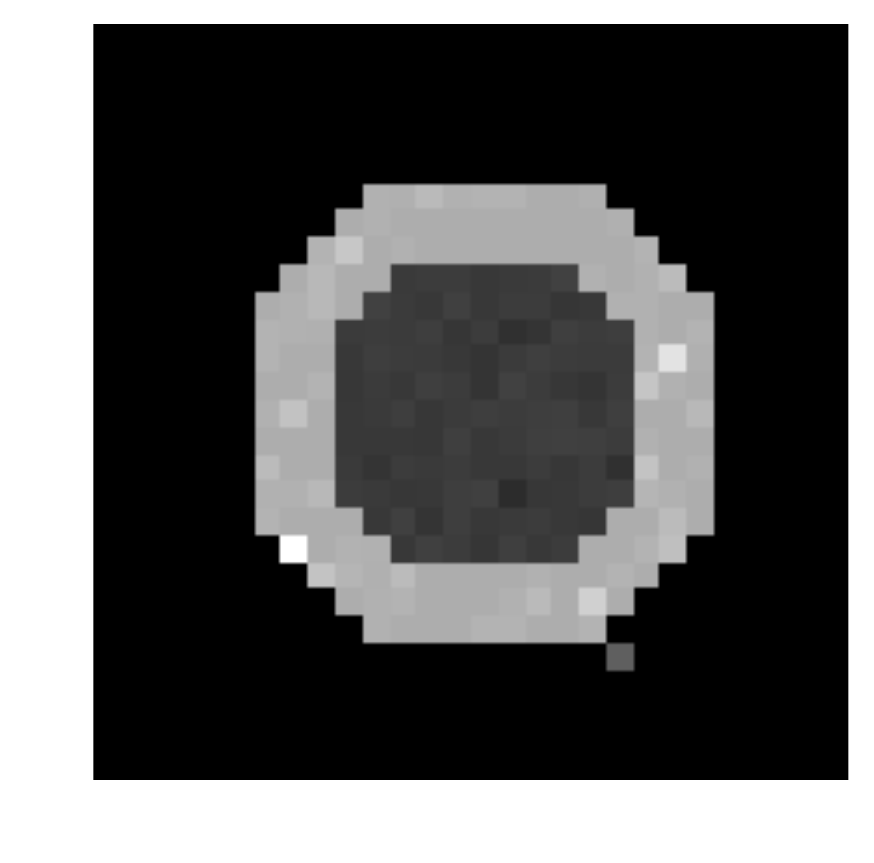}
&
\includegraphics[width=0.1\textwidth]{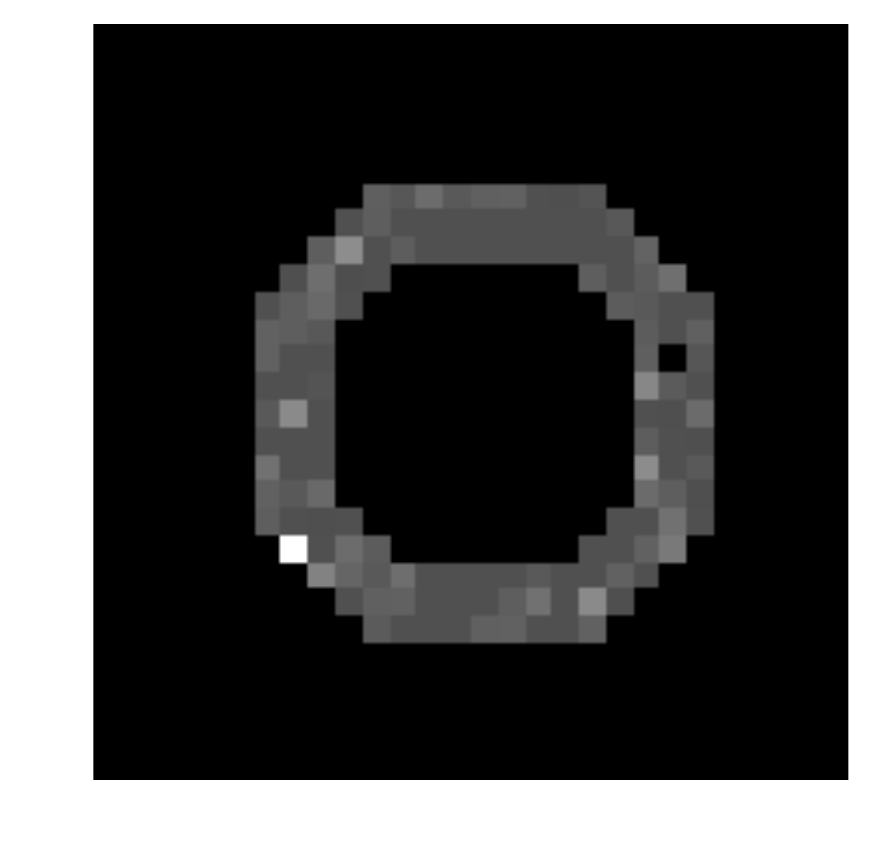}
&
\includegraphics[width=0.1\textwidth]{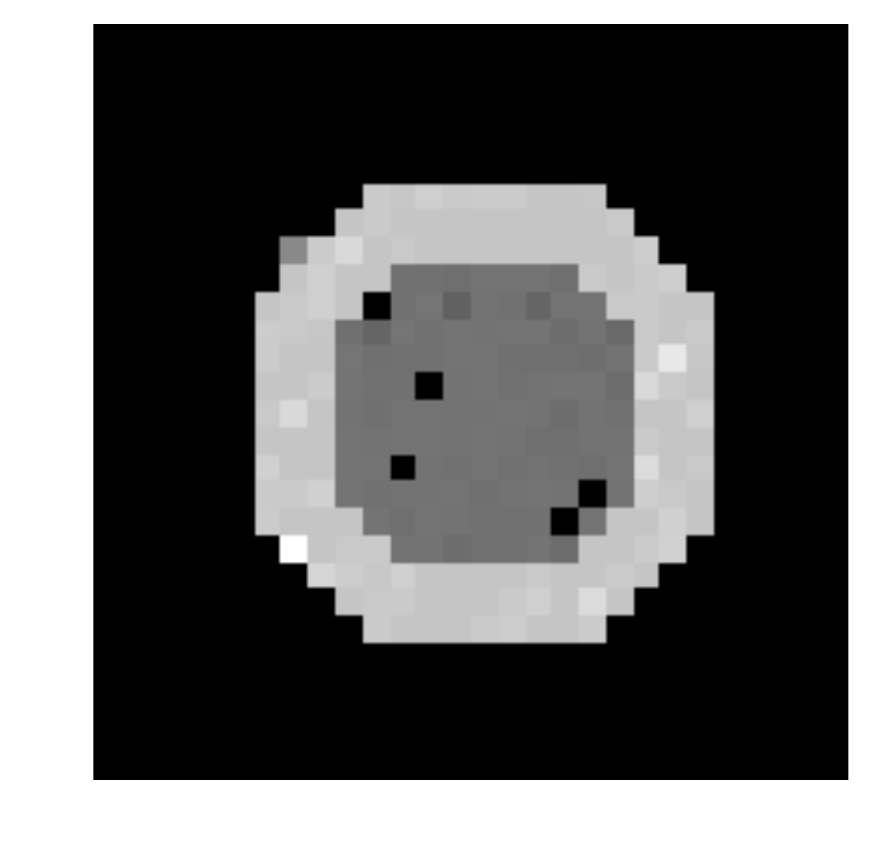}
&
\includegraphics[width=0.1\textwidth]{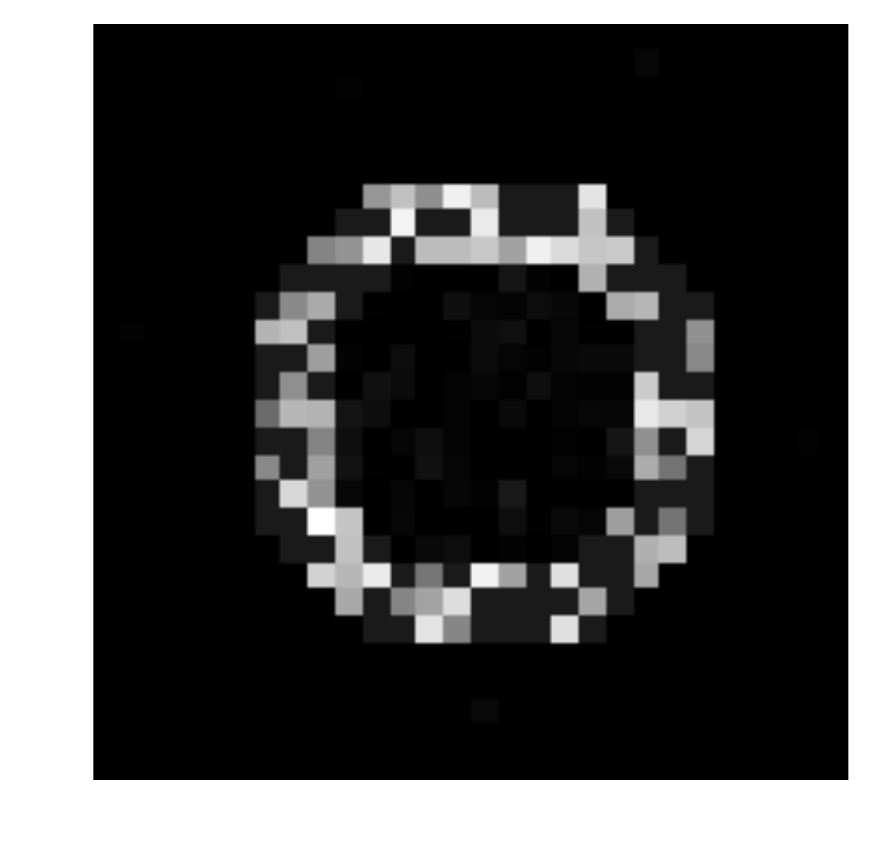}
% &
% \includegraphics[width=0.1\textwidth]{figures/hitmissSofthm-hit_0.pdf}
 \\
 & 
 \copyrightbox[l]{
\includegraphics[width=0.1\textwidth]{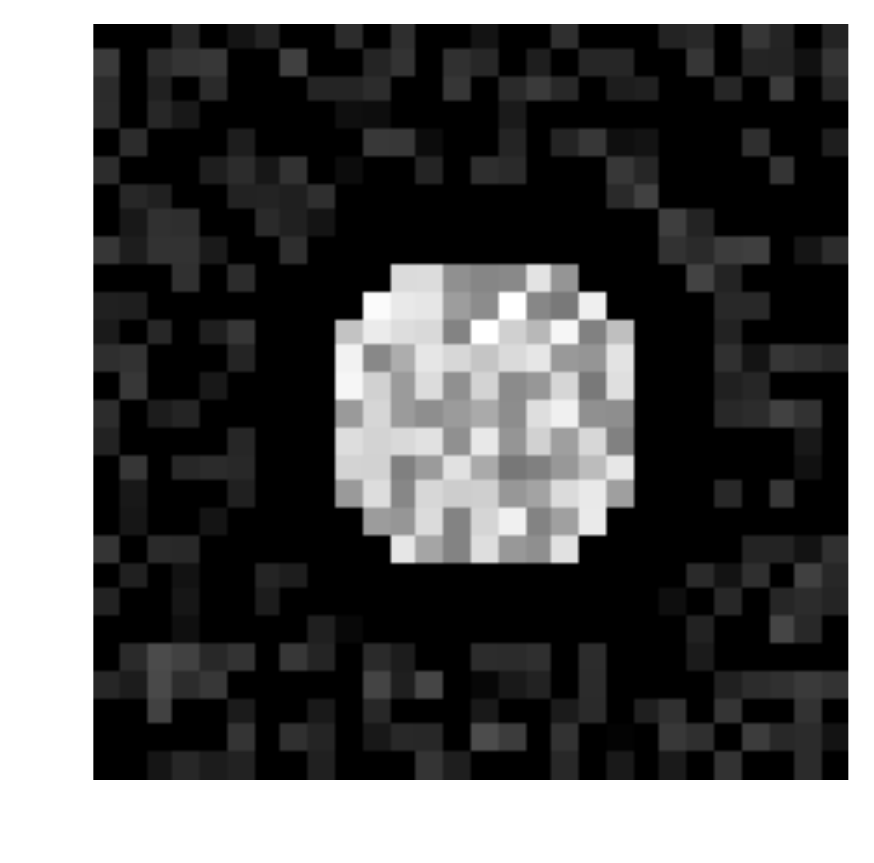}
}{Miss SE}
&
\includegraphics[width=0.1\textwidth]{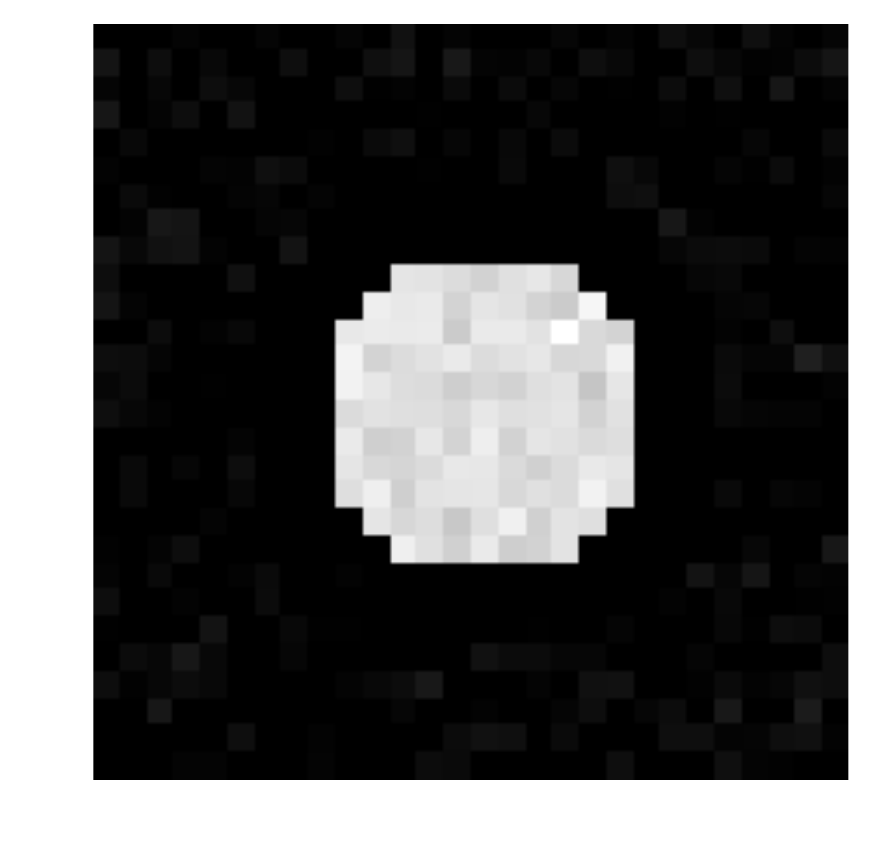}
&
\includegraphics[width=0.1\textwidth]{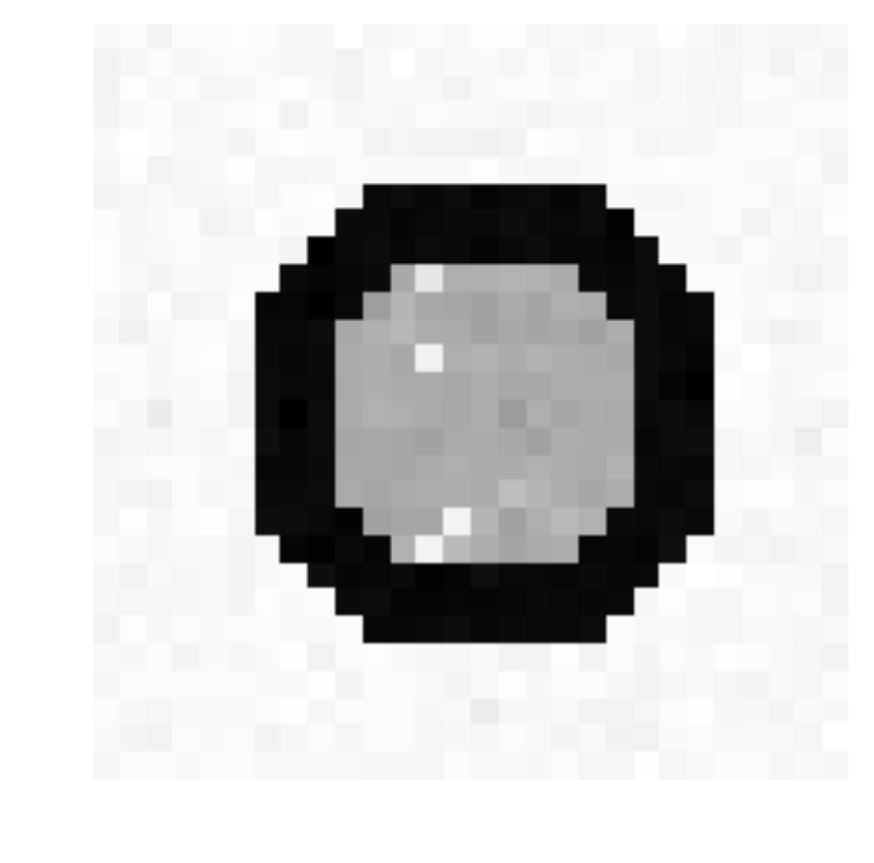}
& 
\includegraphics[width=0.1\textwidth]{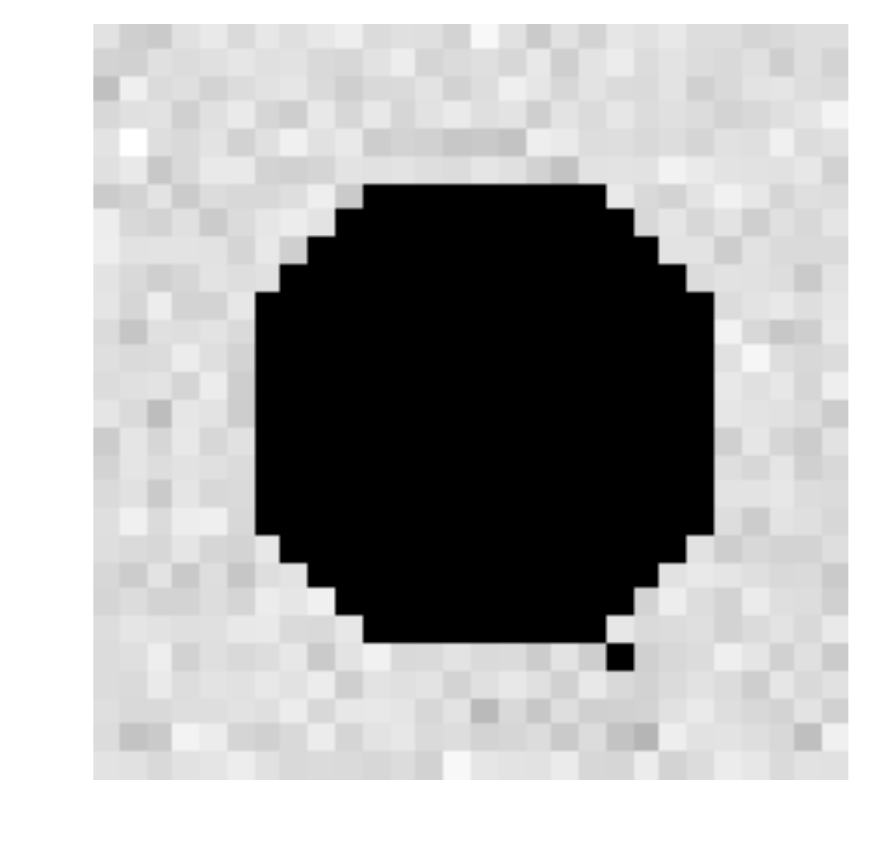}
&
\includegraphics[width=0.1\textwidth]{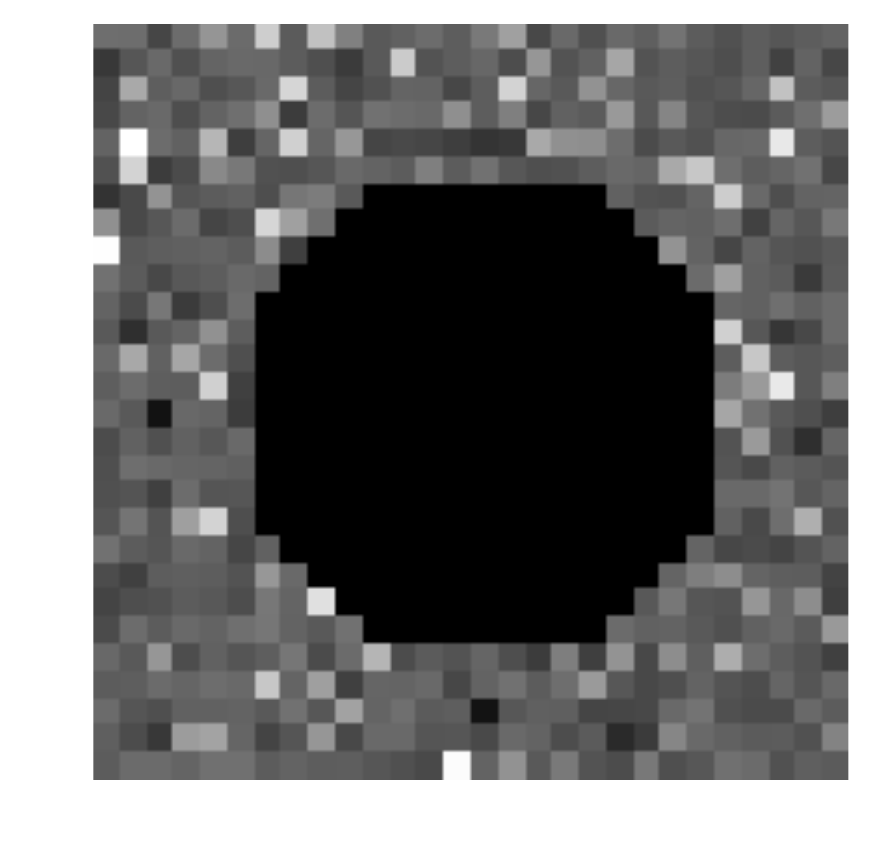}
&
\includegraphics[width=0.1\textwidth]{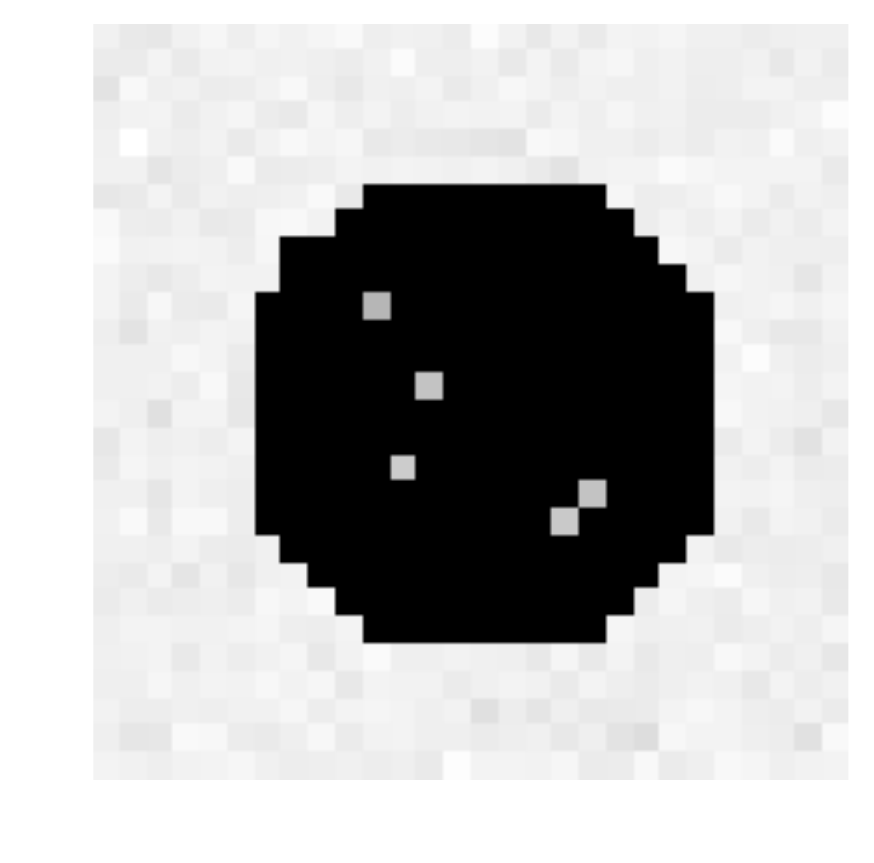}
&
\includegraphics[width=0.1\textwidth]{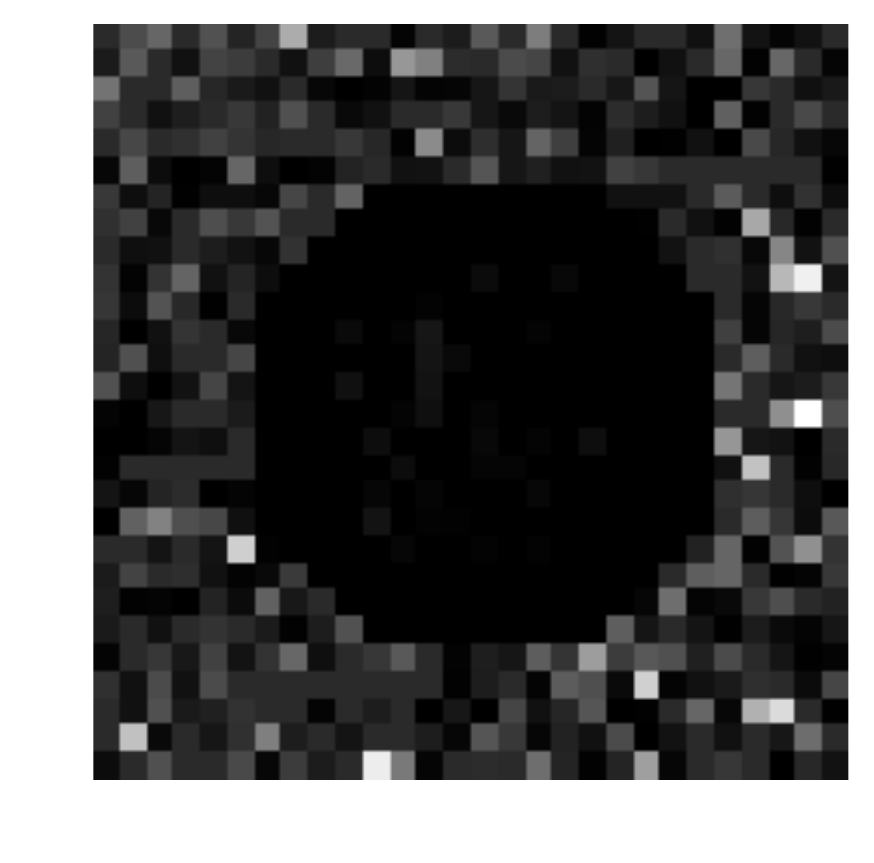}
% &
% \includegraphics[width=0.1\textwidth]{figures/hitmissSofthm-miss_0.pdf}
\\
 (a) Class object & \quad (b) Convolution & (c) Convolution+ReLU & (d) Standard hit-or-miss & (e) hit-or-miss with non-intersecting & (f) hit-or-miss with DNC & (g) hit-or-miss with non-intersecting and DNC & (h) Single SE hit-or-miss
%  & (h) Soft hit-or-miss ($\alpha=0$)
 \\
\end{tabular}
    \caption{Visualization of the learned SEs and filters for synthetic objects. Convolution+ReLU learns the shape of one object, annular ring, and uses it and its invert to discriminate between the two objects. On the other hand, hit-or-miss learns the shapes of both objects. Speckles within the filters may be relevant as the exact matching is not required to obtain a peak classification accuracy due to the discriminatory nature of learning and can be robust against noise and imperfection in the input images.}
    \label{fig:synthetic}
\end{figure*}

This dataset consists of two objects, a solid circle and an annular ring with a hollow at the center, on a $28\times 28$ grid, as shown in the leftmost column in  Fig.~\ref{fig:synthetic}. Two hundred images from each class were generated by perturbing these images with a Gaussian noise with a standard deviation of $0.03$. A single layer NN with two $28\times 28$ hit-or-miss transform/convolution filters without padding was batch-optimized using gradient descent with a learning rate of $0.01$ and momentum $0.9$ for $1000$ epochs.  The network was initialized with a fixed $0.01$ for Dual SEs transforms; and $-0.01$ and $0.01$ for convolution and Single SE transform. We used the mean of squared error as the loss function. 

Fig.~\ref{fig:synthetic} shows the learned filters/SEs. Since convolution itself is a linear operation, we also included convolution+ReLU to make it non-linear and thus comparable to non-linear hit-or-miss transforms. As we can see, convolution+ReLU learns the shape of only one class, annular ring, with foreground shape, hollowed ring, for the hit, and a solid circle the same size as the hole in the ring for the miss.  The filters for the solid circle class are just the opposite of those for the annular ring. In effect, convolution decides based on whether a ring is present or absent in an input image, acting as a relative measure rather than finding a similarity measure with corresponding object shape.  
Contrast these filters against those SEs for the standard hit-or-miss transform. The learned shapes are now consistent with the class objects, e.g., circle and inverted circle for the hit and miss for solid circle object; and ring and inverted ring for the annular ring object. Enforcing the non-intersecting condition helps to learn the solid circle better and the annular ring worse. Adding DNC makes the filters sparse. The SF hit-or-miss transform yields very sparse SEs, e.g.,  SEs for the solid circle includes some dots close to the center in the hit and on the outer-side in the miss, and are sufficient to detect a solid circle. Note that due to the discriminatory nature of learning, exact matching is not required to obtain a peak classification accuracy. Therefore, the speckles within the SEs/filter may be relevant and can be robust against noise and imperfection in the input. A caveat of sparse SEs is that our model can easily be fooled with inputs artificially crafted or sampled from a distribution different from what the model is trained on.

\subsection{Benchmark datasets}

We first provide a brief description of the datasets used in this experiment. \emph{Fashion-MNIST}: This data set consists of fashion articles images of 10 classes; t-shirt/top, trouser, pullover, dress, coat, sandal, shirt, sneaker, bag, and ankle boot. This dataset is similar to the MNIST in term of number of examples ($70,000$), image size ($28\times 28$), and training-test partition ($60,000$/$10,000$), and number of classes ($10$). \emph{Cifar-10}:
This data set consists of $60000$ $32\times 32$ colour images in $10$ classes, with 6000 instances per class.   The dataset is partitioned into training and test with $50,000$ and $10,000$ examples, respectively.

% Some  examples of MNIST, Fashion MNIST, and Cifar are shown in Fig.~\ref{}.

% a.	Datasets: MNIST, Fashion MNIST, and Cifar

% b.	Architectures: ResNet/inception/Densenet, VGG, and MiniVGG. 
% \subsection{Results}
\subsubsection{Impact of initialization}
We consider three distributions, uniform, normal, and half-normal, with different parameters. We optimize miniVGG with standard hit-or-miss transform for $70$ epochs using Adam optimization \cite{kingma2014adam} with a learning rate of $0.001$ and a batch size of $64$. The best test classification accuracy for each experiment is reported in Table~\ref{tab:init-fashion-mnist}.

As seen in Table~\ref{tab:init-fashion-mnist}, initialization can make a big difference. For example, using a normal instead of uniform distribution  increases the accuracy by $20\%$ on Cifar-10 dataset. Half-normal distribution boosts the performance further by $3.61\%$. This improvement can be explained by the fact that normal distribution has a high density around the mid-point. In contrast, half-normal has a high density at the lower end, thus facilitating sparse optimization as fewer elements will contribute to the error. Adopting the initialization condition put forth in Appendix~\ref{initAnalysisSHM} gives the best result.

\begin{table}[htbp]
  \centering
  \renewcommand{\arraystretch}{1.3}
  \caption{Results for different initialization strategies for standard hit-or-miss transform}
    \begin{tabular}{llcc}
    \toprule
    \multirow{2}{*}{Distribution} &  \multirow{2}{*}{Parameter} & \multicolumn{2}{c}{Datasets} \\
    \cmidrule{3-4}
     &  & Fashion-MNIST & Cifar-10 \\
    \midrule
    \multirow{2}{*}{Uniform} & U(-0.01,0.01) & 89.33 & 32.54 \\
          & U(-1,1) & 90.26 & 44.83\\
          [0.5ex] 
    \multirow{2}{*}{Normal} & N(0,0.1) & 89.13 & 57.12\\
          & N(0,1) & 91.56 & 64.15\\[0.5ex]
    \multirow{2}{*}{Half-normal} & HN(0,1) & 92.4 & 67.76 \\
     & According to Appendix.~\ref{initAnalysisSHM} & 92.48 & 69.57 \\
          \bottomrule
    \end{tabular}%
  \label{tab:init-fashion-mnist}%
\end{table}%

\begin{table}[htbp]
  \centering
  \renewcommand{\arraystretch}{1.5}
  \caption{Results for hit-or-miss transforms and convolution}
    \begin{tabular}{lp{2.0cm}cc}
    \toprule
    Methods & \multicolumn{1}{l}{Constraints} & Fashion-MNIST & Cifar-10 \\
    \midrule
    Hit (Erosion) &       & $90.97$ & $56.33$ \\
    Miss (-Dilation) &       & $88.33$ & $53.10$ \\
    Dual SEs hit-or-miss & None & $93.31$ & $72.49$ \\
          & Non-intersecting & $93.25$  & $72.72$ \\
          & DNC ($th$=0.0) & $93.25$ &  $72.91$\\
          & Non-intersecting + DNC ($th$=0.0) & $93.25$ & $72.72$ \\
    Single SE hit-or-miss &    & $93.09$  & $72.90$ \\
    Convolution &  &$\mathbf{94.60}$ & $\mathbf{87.59}$\\
    \bottomrule
    \end{tabular}%
  \label{tab:hit-miss-variants}%
\end{table}%

\subsubsection{hit-or-miss transform and convolution}
The experiment setup is the same except that $150$ epochs is used versus $70$ in prior experiments.  We used Kaimiing initialization \cite{he2015delving} for convolution. For DNC, we used threshold, $th = 0.0$.

As seen in Table~\ref{tab:hit-miss-variants}, foreground aka hit is a better predictor (56\% accuracy on Cifar-10) than background aka miss (53.1\%).  Standard hit-or-miss transform improves the performance further by $16.16\%$ demonstrating the importance of both foreground and background in object detection. 
However, accuracy remains more-or-less the same for the the proposed method after incorporating the non-intersecting condition and DNC. Several factors affect the performance: (i) adding the non-intersecting condition makes the optimization problem more constrained that weakens its approximation power to learn an arbitrary function, and (ii)  DNC space is discontinuous, where no updating occurs during optimization, limiting its ability to learn proper SEs. 

While the hit-or-miss transform enables learning  interpretable SE, convolution outperforms all variants of hit-or-miss transform. This performance gain by convolution is due in part to its superior ability to approximate an arbitrary function, as stated by the universal approximation theorem.  
So, one can trade-off between interpretability and accuracy and select an appropriate operation appropriate for a task.
% So, one can choose to have explainable SE emplopying hit-or-miss transform with the loss of accuaracy or can opt for convolution to obtain maximum accuracy. 

\begin{table}[htbp]
  \centering
  \renewcommand{\arraystretch}{1.6}
  \caption{Results for extensions of the hit-or-miss transform and convolution}
    \begin{tabular}{p{3.5cm} *{3}{c}}
    \toprule
Method & $\alpha$ & Fashion-MNIST & Cifar-10\\
    \midrule
Dual SEs SHM & $1.0$ & $93.74$ & $77.57$ \\
Dual SEs SHM + non-intersecting & $1.0$ & $93.75$  & $77.39$ \\
Dual SEs SHM + DNC & $1.0$  & $93.64$ & $77.32$\\
Dual SEs SHM + non-intersecting + DNC & $1.0$ & $93.78$  & $77.7$ \\
\multirow{1}{*}{Single SE SHM} & $1.0$ & $93.46$  &  $76.95$\\[0.5ex]
%  \cmidrule{1-4}
%  Convolution &  & $94.6$ & $87.59$\\
\multirow{2}{*}{GC1} & $0.5$ &  $94.28$ & $87.28$ \\
 & $1.0$  & $94.32$  & $87.76$\\
%  & $2.0$ &   & \\
Convolution with sum $\quad$ $\quad$ replaced by softmin (1st term of Eq.~(\ref{eq:GC2})) & $0.5$  &  $94.36$ & $87.59$ \\
Convolution with sum $\quad$ $\quad$ replaced by softmax (2nd term of Eq.~(\ref{eq:GC2})) & $0.5$ & $94.44$   & $86.49$ \\
 \multirow{2}{*}{GC2} & $0.5$ & $94.58$ & $\mathbf{88.29}$\\
 & $1.0$ & $\mathbf{94.66}$  & $87.71$\\
%  & $2.0$ &   & \\
    \bottomrule
    \end{tabular}%
  \label{tab:results-miniVGG}%
\end{table}%

% \begin{table}[htbp]
%   \centering
%   \renewcommand{\arraystretch}{1.9}
%   \caption{Results for miniVGG on Fashion-MNIST dataset}
%     \begin{tabular}{l *{3}{c}}
%     \toprule
% & $\alpha$ & Initialization variance, $\sigma^2$ & Accuracy\\
%     \midrule
% \multirow{6}{*}{Generalized convolution} & $0$ & $\frac{1}{n}$ &  \\
%  & $0.25$ & $94.24$ & \\
%  & $0.5$ & $\frac{0.7596}{n^{1.0452}}$  &  \\
%  & $1.0$ & $ \frac{0.6957}{n^{1.2533}}$ &  \\
%  & $2.0$ &  $\frac{1.2168}{n^{1.6818}}$ &  \\
%  & $8.0$ &  & \\

% Soft hit-or-miss (SHM) & $0$ & & $94.24$\\
%     \bottomrule
%     \end{tabular}%
%   \label{tab:results-miniVGG}%
% \end{table}%

% \paragraph{Cifar-10}

% \subsubsection{ResNet-18}
% % \paragraph{MNIST}
% \paragraph{Fashion-MNIST}
% \paragraph{Cifar-10}
\subsubsection{SHM and GC}
Table~\ref{tab:results-miniVGG} reports the results for extensions of the hit-or-miss transform and convolution. Relaxing max/min in the hit-or-miss transform with softer average-weighting operator enhances SHM's performance, though still lags behind standard convolution. GC1 results are at the same level as convolution. GC2 leads the scoreboard with an accuracy of $94.66$ for Fashion-MNIST and $88.29$ for Cifar-10. This indicates that extensions in general boost the results, which reach maximum somewhere between $\alpha=0$ and $\pm \infty$. 

All these experiments share a common story that convolution and the hit-or-miss transform come very close in terms of accuracy for simpler classification tasks (Fashion-MNIST) but the gap becomes wider for challenging tasks with complex objects (Cifar-10). There are many factors behind this performance gap, however the primary reasons can be attributed to (i) its underlying theory of measuring absolute fitness, which enables learning explainable filters but works as a hindrance in achieving top performance, and (ii) the difficulty of optimization with DNC.

\section{Conclusion and Future Work} 
\label{sec:conclusion}
In this article, we provide an in-depth analysis of the theory of grayscale morphology relative to deep neural networks, shedding some critical insights into its limitations and strengths. We also explore an application of a morphological operation, the hit-or-miss transform, that takes into account both foreground and background in measuring the fitness of a target pattern in an image. Unlike binary morphology, conventional grayscale morphological operations consider all pixels regardless of their relevance to a target shape. Furthermore, hit and miss SEs should semantically be non-intersecting. We provide an optimization friendly neural network-based hit-or-miss transform that takes these properties into account.

Specifically, we outline an optimization problem to appropriately learn semantically meaningful and interpretable SEs. Following this formulation, we provide two algorithms for the hit-or-miss transform with one and two SEs. Since max and min in the hit-or-miss equation are too restrictive and overly sensitive to variation and fluctuation in inputs, we relaxed these operators with a parametric generalized mean, yielding a flexible and more powerful transform that leads to better classification accuracy. In the same spirit, we also extend convolution, which outperforms standard convolution on benchmark datasets.

Our analysis and experimental results show that both the hit-or-miss transform and convolution consider both background and foreground, however they differ in the respect that the former provides an absolute measure while the latter gives a relative measure. These differences impact their ability in terms of interpretability and robustness. As better interpretability comes from an absolute measure, morphology leads convolution in this regard. On the other hand, relative measures are more roubust, so convolution outperforms morphology in classification accuracy. Last, quantitative experiments were presented that demonstrate the numeric potential of these networks and qualitative results were demonstrated related to a single hit-or-miss layer.

We limit the focus of the current article to applying morphological operation in deep learning. In the future, we will study how to explain a morphological neural network solely based on the SEs, leveraging the shapes learned by them. Furthermore, we will study how to better handle the discontinuity for DNC. Specifically we will explore other optimization techniques such as genetic algorithms (not stochastic gradient descent-based) with better constraints handling mechanism that will be able to update elements in a bidirectional manner across disjointed spaces. Next, the initialization criteria developed herein was based on curve-fitting and simplified analysis. A future research direction can be toward conducting rigorous mathematical analysis to find exact closed-form equations for variances and co-variances involving generalized mean to enhance the performance further. Last, our qualitative visualization of shape is currently only applicable to a single morphological layer. In future work, we will extend this analysis to multi-layer propagation of morphology to extract explicit shape descriptors for purposes like explainable deep neural shape analysis.
\appendix

% \begin{table*}[htbp]
%   \centering
%   \renewcommand{\arraystretch}{1.6}
%   \caption{Variance of the smooth-max function, $s_\alpha$ vs. $\alpha$}
%     \begin{tabular}{l *{9}{c}}
%     \toprule
%     $\alpha$ &  $0$ & $\pm 0.25$ & $\pm 0.5$ &  $\pm 0.75$ & $\pm 1$ & $\pm 2$ & $\pm 4$ & $\pm 8$ & $\pm \infty$ (max/min)\\
%     \midrule
%     $\sigma_{s_\alpha}^2 / \sigma_x^2$ & $\frac{1}{n}$ & $\frac{1.0945}{n^{0.9933}}$ & $\frac{1.3165}{n^{0.9548}}$ & $\frac{1.4712}{n^{0.8698}}$ & $\frac{1.4374}{n^{0.7467}}$ & $\frac{0.8218}{n^{0.3182}}$ & $\frac{0.5585 }{n^{0.1695}}$ & $\frac{0.5789}{n^{0.2164}}$ & $\frac{0.5976}{n^{0.2427}}$ \\
%     % $\frac{0.5844}{n^{0.2392}}$
%     % complementary variance =  0.6826
%     % $\sigma_{s_\alpha}^2 / \sigma_x^2$ & $\frac{1}{n}$ &  & $\frac{1.0662}{n^{0.8630}}$  &  &  &  &  &  &  $\frac{0.8896}{n^{0.4071}}$\\    
%     \bottomrule
%     \end{tabular}%
%   \label{tab:generalized-mean-distr}%
% \end{table*}%

\subsection{GC2}
Consider a NN layer  consisting of GC2,
\[ y = f  *^{g2} w = n \left(s_{smax,\alpha_1} (f w) + s_{smin, \alpha_2}( f w)\right),\]
followed by Relu activation function
\[z=max(y,0).\]
Let $\sigma_f^2$ and $\sigma_w^2$ be the variances of $f$ and $w$, respectively. Ignoring the covariance between two terms, the variance of the output $y$ will approximately be 
\[\sigma_y^2  \approx n^2 \sigma_{s_{smax,\alpha_1}}'^2 \sigma_f^2 \sigma_w^2 + \sigma_{s_{smmin,\alpha_2}}'^2 \sigma_f^2 \sigma_w^2\]
If we use symmetric soft-max and soft-min, function, then $\alpha_1 = \alpha_2 = \alpha$ and $\sigma_{s_{smax,\alpha_1}}^2  = \sigma_{s_{smax,\alpha_2}}^2 = \sigma_{s_{\alpha}}^2$. This gives
\[\sigma_y^2 \approx 2n^2 \sigma_f^2 \sigma_w^2 \sigma_{s_{\alpha}}'^2.\]
Since $\sigma_z^2 = 0.5 \sigma_y^2$ as shown in \cite{he2015delving} for a symmetric distribution of $y$, As a result
% Since the variance of RelU(x) is $0.5 \sigma_y^2 $, then the variance of the output of ReLu will be
\[\sigma_z^2 \approx n^2 \sigma_f^2 \sigma_w^2 \sigma_{s_{\alpha}}'^2.\]
The output variance $\sigma_z^2$ will be the same as $\sigma_f^2$ if
\[ \sigma_w^2 \approx \frac{1}{n^2\sigma_{s_{\alpha}}'^2},\]
which gives us the variance to initialize the filter weights. GC1 is also initialized with this same variance, which we found to give better results.

\subsection{Soft hit-or-miss transform}
\label{initAnalysisSHM}
Consider a NN layer consisting of softer extension of standard hit and miss transform,
\[f \odot^s (h,m) = s_{smin,\alpha} (f - h) - s_{smin,\alpha}(f + m),\]
followed by Relu activation function
\[z=max(y,0).\]

Let $\sigma_h = \sigma_m$. Then $\sigma_z^2 \approx \sigma_{s,\alpha}'^2 (\sigma_f^2 + \sigma_h^2).$
The condition for $\sigma_z^2$ to be equal to $\sigma_f^2$ is
\[\sigma_h^2 = \sigma_m^2 \approx \left(\frac{1}{\sigma_{s,\alpha}'^2 } - 1\right) \sigma_f^2.\]

If initialized with half-normal distribution, then the variance will be,
\[\sigma_h^2 = \sigma_m^2 \approx \frac{1}{\sigma_{hn}^2}\left(\frac{1}{\sigma_{s,\alpha}'^2 } - 1\right) \sigma_f^2,\]
% Since the variance of half-normal distribution is $(1-2/pi)\sigma^2$, 
where $\sigma_{hn}$ is the ratio of half-normal to normal variances, $\sigma_{hn}^2   = (1 - 2/\pi).$

We use this variance to initialize both standard and proposed hit-or-miss transforms. For $|\alpha|<\infty$, the variance obtained using this equation is very high, causing exploding gradient. To alleviate this, we  scale the hit-or-miss transform equation with $\sigma_{s,\infty}/\sigma_{s,\alpha}$ and initialize SEs the variance for $\alpha=\pm \infty$. 
% \bibliographystyle{abbrv} 
% \makeatletter
% \renewcommand\@biblabel[1]{#1.}
% \makeatother
% DEREK \begingroup
% DEREK \let\itshape\upshape
% DEREK \printbibliography
% DEREK \endgroup
\bibliographystyle{IEEEtran}
\bibliography{IEEEabrv,refs}

% Generated by IEEEtran.bst, version: 1.14 (2015/08/26)
\begin{thebibliography}{10}
\providecommand{\url}[1]{#1}
\csname url@samestyle\endcsname
\providecommand{\newblock}{\relax}
\providecommand{\bibinfo}[2]{#2}
\providecommand{\BIBentrySTDinterwordspacing}{\spaceskip=0pt\relax}
\providecommand{\BIBentryALTinterwordstretchfactor}{4}
\providecommand{\BIBentryALTinterwordspacing}{\spaceskip=\fontdimen2\font plus
\BIBentryALTinterwordstretchfactor\fontdimen3\font minus
  \fontdimen4\font\relax}
\providecommand{\BIBforeignlanguage}[2]{{%
\expandafter\ifx\csname l@#1\endcsname\relax
\typeout{** WARNING: IEEEtran.bst: No hyphenation pattern has been}%
\typeout{** loaded for the language `#1'. Using the pattern for}%
\typeout{** the default language instead.}%
\else
\language=\csname l@#1\endcsname
\fi
#2}}
\providecommand{\BIBdecl}{\relax}
\BIBdecl

\bibitem{geirhos2018imagenettrained}
R.~Geirhos, P.~Rubisch, C.~Michaelis, M.~Bethge, F.~A. Wichmann, and
  W.~Brendel, ``Imagenet-trained cnns are biased towards texture; increasing
  shape bias improves accuracy and robustness,'' 2018.

\bibitem{springenberg2014striving}
J.~T. Springenberg, A.~Dosovitskiy, T.~Brox, and M.~Riedmiller, ``Striving for
  simplicity: The all convolutional net,'' \emph{arXiv preprint
  arXiv:1412.6806}, 2014.

\bibitem{simonyan2013deep}
K.~Simonyan, A.~Vedaldi, and A.~Zisserman, ``Deep inside convolutional
  networks: Visualising image classification models and saliency maps,''
  \emph{arXiv preprint arXiv:1312.6034}, 2013.

\bibitem{mellouli2019morphological}
D.~Mellouli, T.~M. Hamdani, J.~J. Sanchez-Medina, M.~B. Ayed, and A.~M. Alimi,
  ``Morphological convolutional neural network architecture for digit
  recognition,'' \emph{IEEE transactions on neural networks and learning
  systems}, 2019.

\bibitem{halkiotis2007automatic}
S.~Halkiotis, T.~Botsis, and M.~Rangoussi, ``Automatic detection of clustered
  microcalcifications in digital mammograms using mathematical morphology and
  neural networks,'' \emph{Signal Processing}, vol.~87, no.~7, pp. 1559--1568,
  2007.

\bibitem{won1995morphological}
Y.~Won and P.~D. Gader, ``Morphological shared-weight neural network for
  pattern classification and automatic target detection,'' in \emph{Proceedings
  of ICNN'95-International Conference on Neural Networks}, vol.~4.\hskip 1em
  plus 0.5em minus 0.4em\relax IEEE, 1995, pp. 2134--2138.

\bibitem{won1997morphological}
Y.~Won, P.~D. Gader, and P.~C. Coffield, ``Morphological shared-weight networks
  with applications to automatic target recognition,'' \emph{IEEE Transactions
  on neural networks}, vol.~8, no.~5, pp. 1195--1203, 1997.

\bibitem{zheng2006morphological}
H.~Zheng, L.~Pan, and L.~Li, ``A morphological neural network approach for
  vehicle detection from high resolution satellite imagery,'' in
  \emph{International Conference on Neural Information Processing}.\hskip 1em
  plus 0.5em minus 0.4em\relax Springer, 2006, pp. 99--106.

\bibitem{sulehria2008vehicle}
H.~K. Sulehria, D.~I. Ye~Zhang, and A.~K. Sulehria, ``Vehicle number plate
  recognition using mathematical morphology and neural networks,'' \emph{WSEAS
  Transactions on Computers}, vol.~7, no.~6, pp. 781--790, 2008.

\bibitem{jin2007vehicle}
X.~Jin and C.~H. Davis, ``Vehicle detection from high-resolution satellite
  imagery using morphological shared-weight neural networks,'' \emph{Image and
  Vision Computing}, vol.~25, no.~9, pp. 1422--1431, 2007.

\bibitem{raducanu2001morphological}
B.~Raducanu, M.~Grana, and P.~Sussner, ``Morphological neural networks for
  vision based self-localization,'' in \emph{Proceedings 2001 ICRA. IEEE
  International Conference on Robotics and Automation (Cat. No. 01CH37164)},
  vol.~2.\hskip 1em plus 0.5em minus 0.4em\relax IEEE, 2001, pp. 2059--2064.

\bibitem{gader2000morphological}
P.~D. Gader, M.~A. Khabou, and A.~Koldobsky, ``Morphological regularization
  neural networks,'' \emph{Pattern Recognition}, vol.~33, no.~6, pp. 935--944,
  2000.

\bibitem{hocaoglu2003domain}
A.~K. Hocaoglu and P.~D. Gader, ``Domain learning using choquet integral-based
  morphological shared weight neural networks,'' \emph{Image and Vision
  Computing}, vol.~21, no.~7, pp. 663--673, 2003.

\bibitem{khabou2000ladar}
M.~A. Khabou, P.~D. Gader, and J.~M. Keller, ``Ladar target detection using
  morphological shared-weight neural networks,'' \emph{Machine Vision and
  Applications}, vol.~11, no.~6, pp. 300--305, 2000.

\bibitem{theera1998detection}
N.~Theera-Umpon, M.~A. Khabou, P.~D. Gader, J.~M. Keller, H.~Shi, and H.~Li,
  ``Detection and classification of mstar objects via morphological
  shared-weight neural networks,'' in \emph{Algorithms for Synthetic Aperture
  Radar Imagery V}, vol. 3370.\hskip 1em plus 0.5em minus 0.4em\relax
  International Society for Optics and Photonics, 1998, pp. 530--540.

\bibitem{ouadou2017vehicle}
A.~Ouadou, ``Vehicle detection using morphological shared-weight neural network
  in the multiple instance learning framework,'' Ph.D. dissertation, University
  of Missouri--Columbia, 2017.

\bibitem{perret2009robust}
B.~Perret, S.~Lef{\`e}vre, and C.~Collet, ``A robust hit-or-miss transform for
  template matching applied to very noisy astronomical images,'' \emph{Pattern
  Recognition}, vol.~42, no.~11, pp. 2470--2480, 2009.

\bibitem{chatzis2000generalized}
V.~Chatzis and I.~Pitas, ``A generalized fuzzy mathematical morphology and its
  application in robust 2-d and 3-d object representation,'' \emph{IEEE
  Transactions on Image Processing}, vol.~9, no.~10, pp. 1798--1810, 2000.

\bibitem{ta2010nonlocal}
V.-T. Ta, A.~Elmoataz, and O.~L{\'e}zoray, ``Nonlocal pdes-based morphology on
  weighted graphs for image and data processing,'' \emph{IEEE transactions on
  Image Processing}, vol.~20, no.~6, pp. 1504--1516, 2010.

\bibitem{bouaynaya2008theoretical}
N.~Bouaynaya and D.~Schonfeld, ``Theoretical foundations of spatially-variant
  mathematical morphology part ii: Gray-level images,'' \emph{IEEE Transactions
  on pattern analysis and machine intelligence}, vol.~30, no.~5, pp. 837--850,
  2008.

\bibitem{ji1992fast}
L.~Ji and J.~Piper, ``Fast homotopy-preserving skeletons using mathematical
  morphology,'' \emph{IEEE Transactions on Pattern Analysis \& Machine
  Intelligence}, no.~6, pp. 653--664, 1992.

\bibitem{zana2001segmentation}
F.~Zana and J.-C. Klein, ``Segmentation of vessel-like patterns using
  mathematical morphology and curvature evaluation,'' \emph{IEEE transactions
  on image processing}, vol.~10, no.~7, pp. 1010--1019, 2001.

\bibitem{urbach2007connected}
E.~R. Urbach, J.~B. Roerdink, and M.~H. Wilkinson, ``Connected shape-size
  pattern spectra for rotation and scale-invariant classification of gray-scale
  images,'' \emph{IEEE Transactions on Pattern Analysis and Machine
  Intelligence}, vol.~29, no.~2, pp. 272--285, 2007.

\bibitem{palmer1997locating}
P.~L. Palmer and M.~Petrou, ``Locating boundaries of textured regions,''
  \emph{IEEE transactions on geoscience and remote sensing}, vol.~35, no.~5,
  pp. 1367--1371, 1997.

\bibitem{haralick1987image}
R.~M. Haralick, S.~R. Sternberg, and X.~Zhuang, ``Image analysis using
  mathematical morphology,'' \emph{IEEE transactions on pattern analysis and
  machine intelligence}, no.~4, pp. 532--550, 1987.

\bibitem{sinha1992fuzzy}
D.~Sinha and E.~R. Dougherty, ``Fuzzy mathematical morphology,'' \emph{Journal
  of Visual Communication and Image Representation}, vol.~3, no.~3, pp.
  286--302, 1992.

\bibitem{nogueira2019introduction}
K.~Nogueira, J.~Chanussot, M.~D. Mura, W.~R. Schwartz, and J.~A.~d. Santos,
  ``An introduction to deep morphological networks,'' \emph{arXiv preprint
  arXiv:1906.01751}, 2019.

\bibitem{gader1994image}
P.~D. Gader, Y.~Won, and M.~A. Khabou, ``Image algebra networks for pattern
  classification,'' in \emph{Image Algebra and Morphological Image Processing
  V}, vol. 2300.\hskip 1em plus 0.5em minus 0.4em\relax International Society
  for Optics and Photonics, 1994, pp. 157--168.

\bibitem{khabou1999morphological}
M.~A. Khabou, P.~D. Gader, and J.~M. Keller, ``Morphological shared-weight
  neural networks: A tool for automatic target recognition beyond the visible
  spectrum,'' in \emph{Proceedings IEEE Workshop on Computer Vision Beyond the
  Visible Spectrum: Methods and Applications (CVBVS'99)}.\hskip 1em plus 0.5em
  minus 0.4em\relax IEEE, 1999, pp. 101--109.

\bibitem{dougherty1992introduction}
E.~R. Dougherty, ``An introduction to morphological image processing,''
  \emph{SPIE, 1992}, 1992.

\bibitem{gonzalez2002digital}
R.~C. Gonzalez, R.~E. Woods \emph{et~al.}, ``Digital image processing,'' 2002.

\bibitem{le1990constrained}
Y.~LE~CUN, ``Constrained neural networks for unicon-strained handwritten digit
  recognition,'' \emph{Proc. Fronties in Handwritting Recognition}, pp.
  145--151, 1990.

\bibitem{590053}
P.~{Maragos}, ``Optimal morphological approaches to image matching and object
  detection,'' in \emph{[1988 Proceedings] Second International Conference on
  Computer Vision}, 1988, pp. 695--699.

\bibitem{he2015delving}
K.~He, X.~Zhang, S.~Ren, and J.~Sun, ``Delving deep into rectifiers: Surpassing
  human-level performance on imagenet classification,'' in \emph{Proceedings of
  the IEEE international conference on computer vision}, 2015, pp. 1026--1034.

\bibitem{simonyan2014very}
K.~Simonyan and A.~Zisserman, ``Very deep convolutional networks for
  large-scale image recognition,'' \emph{arXiv preprint arXiv:1409.1556}, 2014.

\bibitem{kingma2014adam}
D.~Kingma and J.~Ba, ``Adam: A method for stochastic optimization,'' in
  \emph{3rd International Conference for Learning Representations}, 2015.

\end{thebibliography}

\begin{IEEEbiography}[{\includegraphics[width=1in,height=1.25in,clip,keepaspectratio]{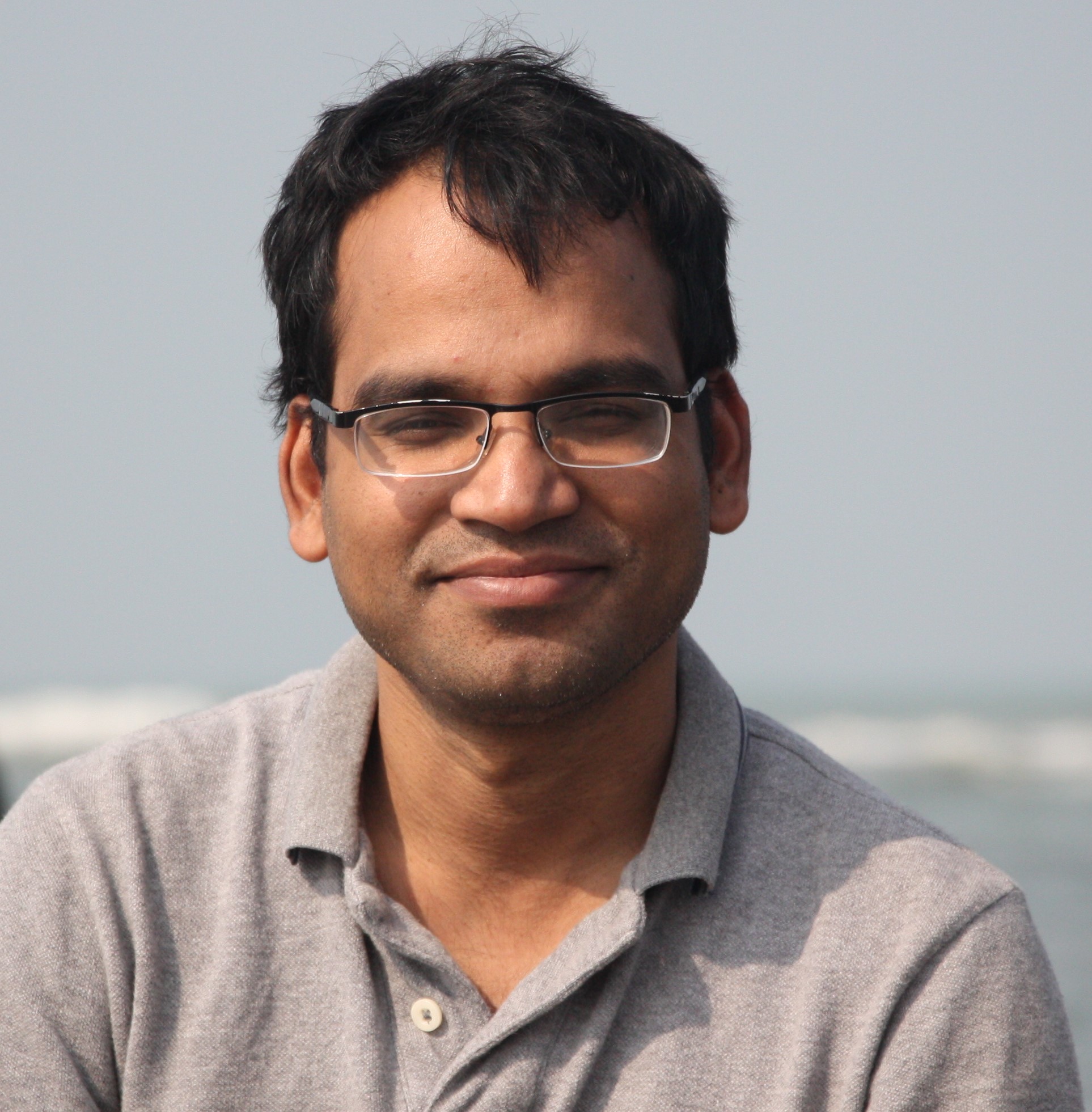}}]{Muhammad Aminul Islam} (M'18) received the B.Sc. degree in Electrical and Electronic Engineering from Bangladesh University of Engineering and Technology, Dhaka, Bangladesh, in 2005 and the Ph.D. degree in Electrical and Computer Engineering from Mississippi State University, Starkville, MS, USA in 2018.

He is currently an Assistant Professor in the Department of Electrical \& Computer Engineering and Computer Science at the University of New Haven (UNH). His research interests include deep learning, computer vision, information fusion, autonomous driving, and remote sensing.
% \begin{IEEEbiography}{Michael Shell}
% Biography text here.
\end{IEEEbiography}

\begin{IEEEbiography}[{\includegraphics[width=1in,height=1.25in]{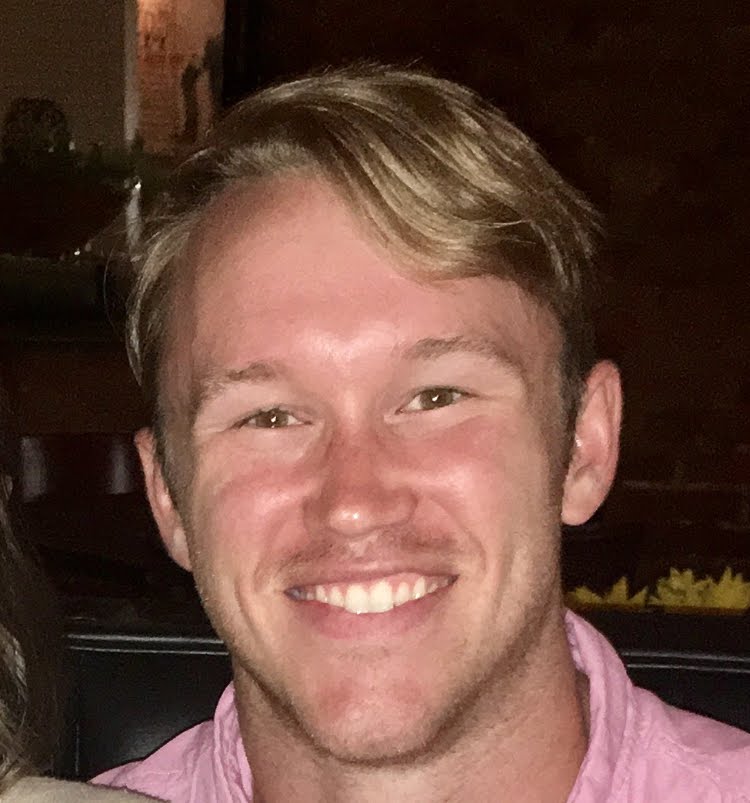}}]{Bryce Murray} received his B.S. in Computer Science, Mathematics, and Physics from Mississippi College, Clinton, MS, USA, in 2015. He then received a M.S. in Electrical and Computer Engineering from Mississippi State University, Mississippi State, MS, USA, in 2018. 

He is a Ph.D. candidate at the University of Missouri, Columbia, MO, USA. His interests include data/information fusion, machine learning, deep learning, computer vision, remote sensing, and eXplainable AI. 
\end{IEEEbiography}

\vskip 0pt plus -1fil

\begin{IEEEbiography}[{\includegraphics[width=1in,height=1.25in,clip,keepaspectratio]{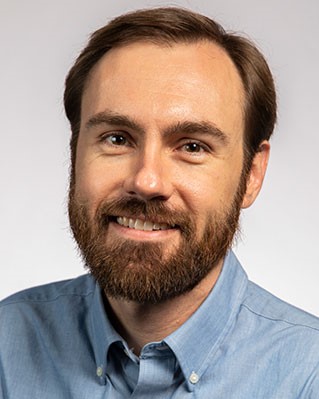}}]{Andrew Buck} (S'11-M'18) received the B.S. degrees in electrical engineering and computer engineering in 2009, M.S. degree in computer engineering in 2012, and Ph.D. in electrical and computer engineering in 2018, all from the University of Missouri, Columbia, MO, USA.

He is an Assistant Research Professor at the University of Missouri in the Electrical Engineering and Computer Science (EECS) department. His research interests include intelligent agents, deep learning, and computer vision.
\end{IEEEbiography}

% \vskip 0pt plus -1fil

\begin{IEEEbiography}[{\includegraphics[width=1in,height=1.25in,clip,keepaspectratio]{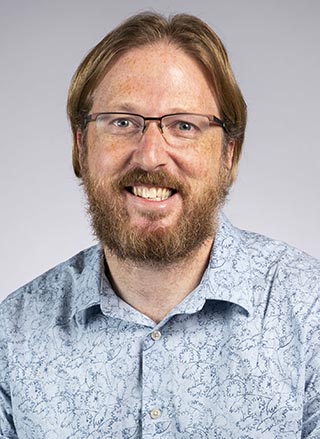}}]{Derek T. Anderson} (SM'13) received the Ph.D. degree in electrical and computer engineering (ECE) from the University of Missouri, Columbia, MO, USA, in 2010. 

He is an Associate Professor in electrical engineering and computer science (EECS) at the University of Missouri and director of the Mizzou Information and Data Fusion Laboratory (MINDFUL). His research is information fusion in computational intelligence for signal/image processing, computer vision, and geospatial applications. Dr. Anderson has published over a 150 articles. He received the best overall paper award at the 2012 IEEE International Conference on Fuzzy Systems (FUZZIEEE), and the 2008 FUZZ-IEEE best student paper award. He was the Program Co-Chair of FUZZ-IEEE 2019, an Associate Editor for the IEEE Transactions on Fuzzy Systems, Vice Chair of the IEEE CIS Fuzzy Systems Technical Committee (FSTC), and an Area Editor for the International Journal of Uncertainty, Fuzziness and Knowledge-Based Systems.
\end{IEEEbiography}

% \vskip 0pt plus -1fil

\begin{IEEEbiography}[{\includegraphics[width=1in]{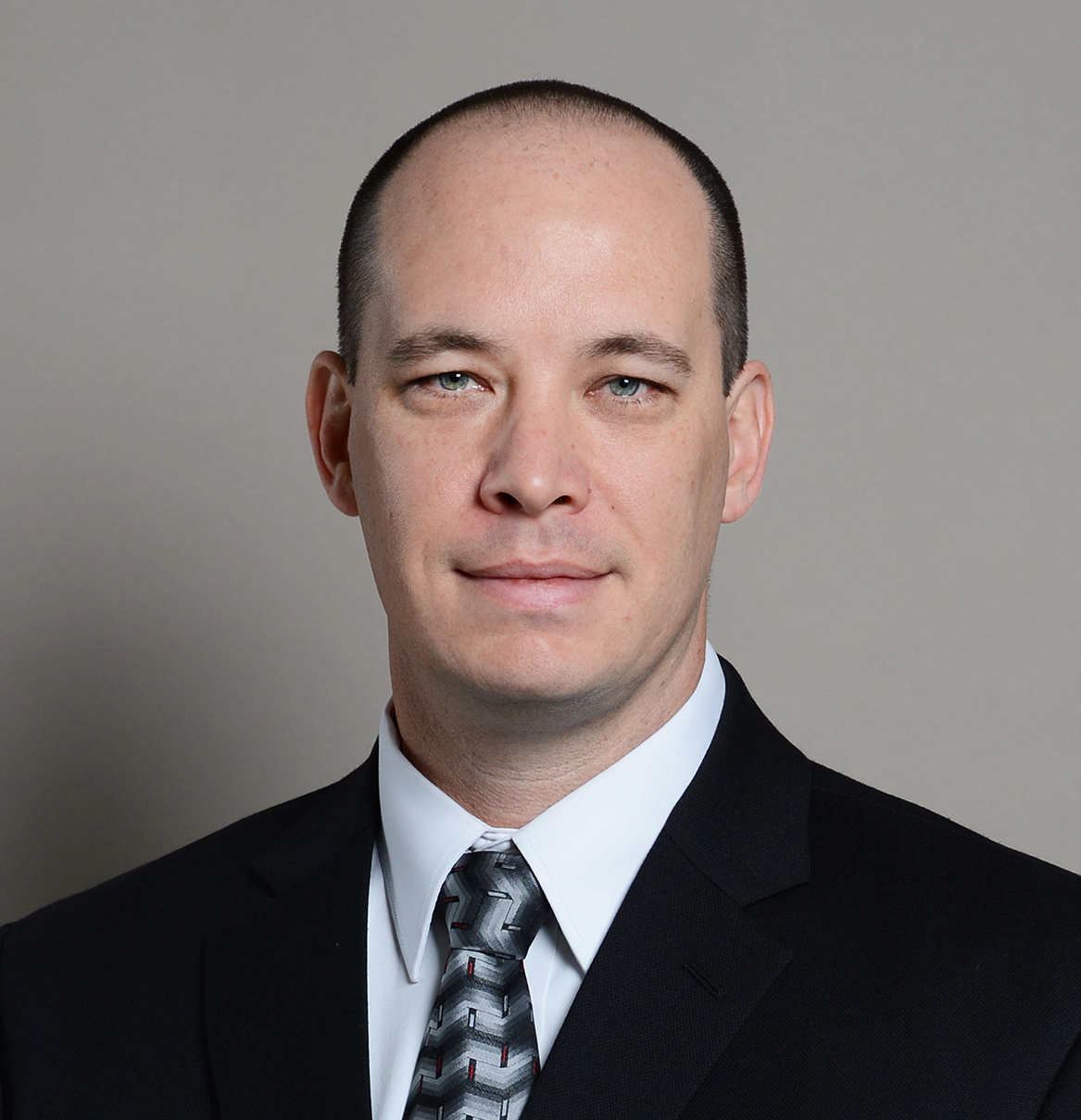}}]{Grant Scott} (S'02-M'09-SM'17) received the B.S. and M.S. degrees in computer science and the
Ph.D. degree in computer engineering and computer science from the University of Missouri, Columbia,
in 2001, 2003, and 2008, respectively. 

He is a founding Director of the Data Science and Analytics Masters Degree program at the University
of Missouri. He is an Assistant Professor with the Electrical Engineering and Computer Science Department, University of Missouri. 

Dr. Scott is exploring novel integrations of computational hardware and software to facilitate high performance advances in large-scale data science, computer vision, and pattern recognition. His current research efforts encompass areas such as: real-time processing of large-scale sensor networks, parallel/distributed systems, and Internet of Things (IoT) data; deep learning technologies applied to geospatial data sets for land cover classification and object recognition;  extensions of enterprise RDBMS with HPC co-processors; crowd-source information mining and multi-modal analytics; high performance \& scalable content-based retrieval (geospatial data, imagery, biomedical); imagery and geospatial data analysis, feature extraction, object-based analysis, and exploitation; pattern recognition databases and knowledge-driven high-dimensional indexing; and image geolocation. He has leveraged this experience in the development of innovative remote sensing (satellite and airborne) change detection technologies, resulting in 5 US Patents. He has participated in a variety of professional networking and academic events, as well as worked with a variety of groups to bring data science training to their people (MU Public Policy, USDA, IEEE international conferences).
\end{IEEEbiography}

\begin{IEEEbiography}[{\includegraphics[width=1in,height=1.25in]{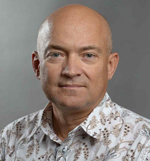}}]{Mihail Popescu}
 received his B.S. degree in Nuclear Engineering from the Bucharest Polytechnic Institute in 1987. Subsequently, he received his M.S. degree in Medical Physics in 1995, his M.S. degree in Electrical and Computer Engineering in 1997, and his Ph.D. degree in Computer Engineering and Computer Science in 2003 from the University of Missouri. He is currently a Professor with the Department of Health Management and Informatics, School of Medicine, at the University of Missouri in Columbia, Missouri, USA. Dr. Popescu is interested in machine learning and medical decision support systems. His current research focus is developing decision support systems for early illness recognition in elderly and investigating sensor data summarization and visualization methodologies. He has authored or coauthored more than 160 technical publications. He is a senior IEEE member.
\end{IEEEbiography}

\vskip 0pt plus -1fil

\begin{IEEEbiography}[{\includegraphics[width=1in,height=1.25in]{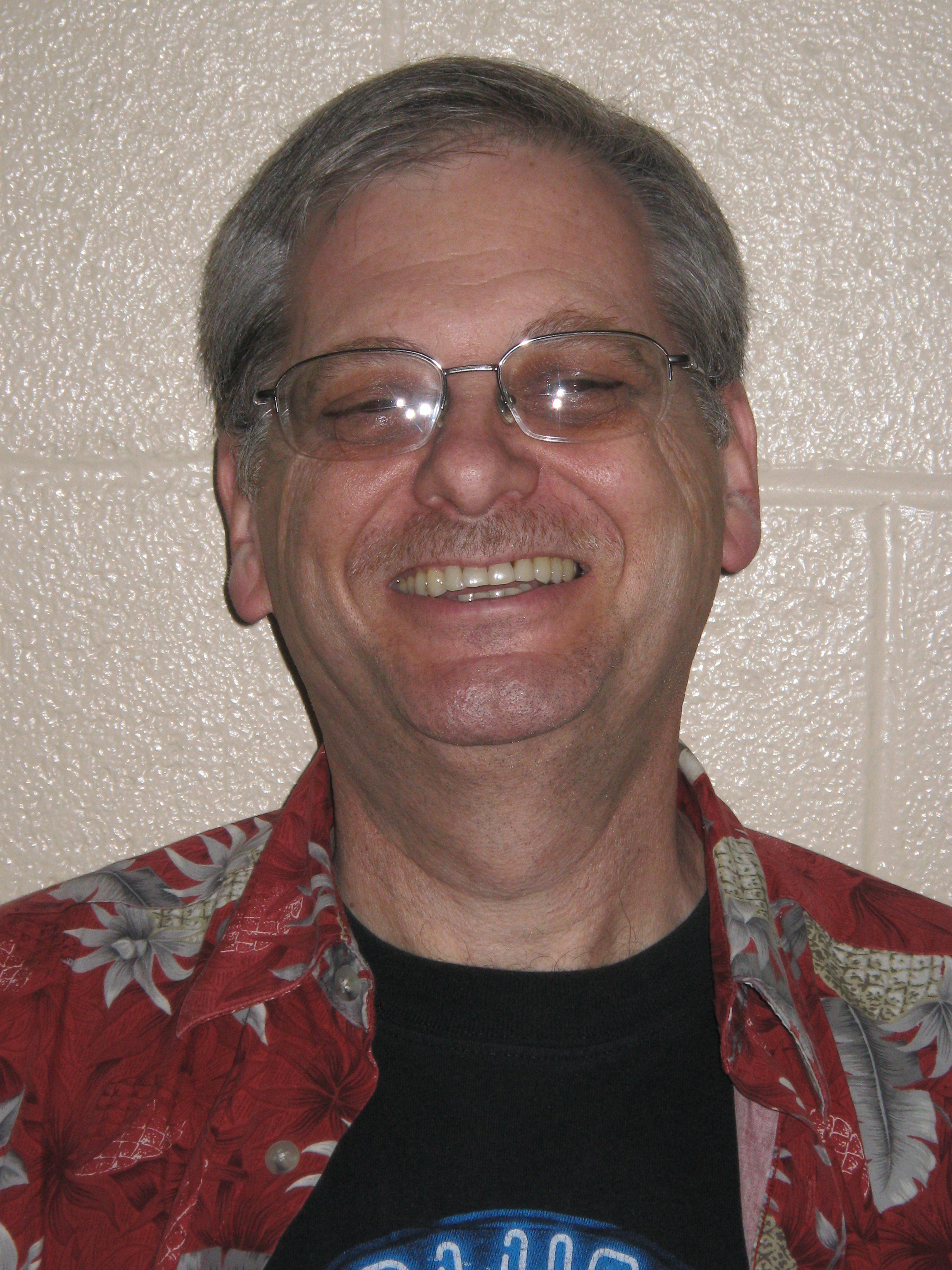}}]{James Keller} is now the University of Missouri Curators Distinguished Professor Emertitus in the Electrical Engineering and Computer Science Department on the Columbia campus. Jim is an Honorary Proferssor at the University of Nottingham. 

His research interests center on computational intelligence with a focus on problems in computer vision, pattern recognition, and information fusion including bioinformatics, spatial reasoning, geospatial intelligence, landmine
detection and technology for eldercare. Professor Keller has been funded by a variety of
government and industry organizations and has coauthored over 500 technical publications.
Jim is a Life Fellow of the IEEE, is an IFSA Fellow, and a past President of NAFIPS. He
received the 2007 Fuzzy Systems Pioneer Award and the 2010 Meritorious Service Award from
the IEEE Computational Intelligence Society. He finished a full six year term as Editor-in-Chief
of the IEEE Transactions on Fuzzy Systems, followed by being the Vice President for
Publications of the IEEE CIS from 2005-2008, and then an elected CIS Adcom member. He is
VP Pubs for CIS again, and has served as the IEEE TAB Transactions Chair and as a member of
the IEEE Publication Review and Advisory Committee from 2010 to 2017. Jim has had many
conference positions and duties over the years. 
\end{IEEEbiography}

% that's all folks
\end{document}